# A Comprehensive Dynamic Simulation Framework for Coupled Neuromusculoskeletal-Exoskeletal Systems

Wei Jin[1,2,#], Jiaqi Liu[3,#], Qiwei Zhang[1,2], Xiaoxu Zhang[1,2], Qining Wang[4], Hongbin Fang[1,2,*], and Jian Xu[3]

[1] Institute of AI and Robotics, State Key Laboratory of Medical Neurobiology, MOE Engineering Research Center of AI & Robotics, Fudan University, Shanghai 200433, China

[2] Yiwu Research Institute, Fudan University, Yiwu, Zhejiang Province 322000, China

[3] School of Aerospace Engineering and Applied Mechanics, Tongji University, Shanghai 200092, China

[4] College of Engineering, Peking University, Beijing 100871, China

[#] W.J. and J.L. contributed equally to this work.

[*] To whom correspondence should be addressed. Email: fanghongbin@fudan.edu.cn (H.F.)

**Abstract:** The modeling and simulation of coupled neuromusculoskeletal-exoskeletal systems play a crucial role in human biomechanical analysis, as well as in the design and control of exoskeletons. However, conventional dynamic simulation frameworks have limitations due to their reliance on experimental data and their inability to capture comprehensive biomechanical signals and dynamic responses. To address these challenges, we introduce an optimization-based dynamic simulation framework that integrates a complete neuromusculoskeletal feedback loop, rigid-body dynamics, human-exoskeleton interaction, and foot-ground contact. Without relying on experimental measurements or empirical data, our framework employs a stepwise optimization process to determine muscle reflex parameters, taking into account multidimensional criteria. This allows the framework to generate a full range of kinematic and biomechanical signals, including muscle activations, muscle forces, joint torques, etc., which are typically challenging to measure experimentally. To validate the effectiveness of the framework, we compare the simulated results with experimental data obtained from a healthy subject wearing an exoskeleton while walking at different speeds (0.9, 1.0, and 1.1 m/s) and terrains (flat and uphill). The results demonstrate that our framework can effectively and accurately capture the qualitative differences in muscle activity associated with different functions, as well as the evolutionary patterns of muscle activity and kinematic signals under varying walking conditions. The simulation framework we propose has the potential to facilitate gait analysis and performance evaluation of coupled human-exoskeleton systems, as well as enable efficient and cost-effective testing of novel exoskeleton designs and control strategies.

# 1. Introduction

Wearable exoskeletons have garnered considerable attention in recent years due to their potential in facilitating human movement-related applications, making them a focal point of biomechanical and robotic research [1]. For example, gait rehabilitation robots have demonstrated superior effectiveness compared to conventional treatments in restoring mobility for patients with spinal cord injuries [2]. Passive exoskeletons have been developed to mitigate the risk of strain-related injuries in the workplace [3] and reduce energy expenditure during movement [4]. On the other hand, powered exoskeletons are designed to provide active assistance in movement for healthy individuals and those with motor impairments [5–8]. Their applications encompass tasks such as aiding individuals in carrying heavy objects over long distances [9], facilitating long-term rehabilitation, supporting the mobility of injured or elderly individuals [10], etc. Despite their immense application potential, the design, control, and performance evaluation of wearable exoskeletons present significant challenges due to several reasons. Firstly, wearable exoskeletons are closely coupled to the human body and therefore necessitate stringent safety requirements, as even minor malfunctions can have severe consequences. Secondly, evaluating the performance of exoskeletons, including aspects like reliability, human-machine interaction, and fault tolerance, etc., demands costly prototyping processes and laborious experiments involving human subjects. Lastly, Finally, the diversity of environmental and task conditions further aggravates the burden of experimental research.

  Given the significant time, labor, and economic costs for experimental studies, conducting dynamic simulations becomes a viable alternative for evaluating exoskeleton performance. Developing a dynamic simulation framework that integrates both the human skeleton and the exoskeleton would enable the estimation of the kinematics of the coupled human-exoskeleton system outside the laboratory setting. This framework facilitates effective analysis of the effects of exoskeletons on human motions and enables convenient assessment of the level of human-exoskeleton interaction. Moreover, dynamic simulations eliminate the need for frequent fabrication of new exoskeleton prototypes and costly experiments, making it possible to test new or improved exoskeleton designs and control strategies at low costs to enhance performance. Additionally, by incorporating the human neuromusculoskeletal system into the dynamic simulation framework, it becomes feasible to perform human gait analysis and obtain valuable information on joint torques, muscle forces, and muscle activations that are challenging to measure experimentally; such information is crucial for analyzing the assistance provided by exoskeletons in human movement. Furthermore, by incorporating tunable human-environment interaction within the dynamic simulation framework, it becomes possible to evaluate the performance of human-exoskeleton coupled systems across different tasks and environments. However, the existing dynamic models and simulation frameworks have not comprehensively addressed all these factors, thereby limiting the widespread application of dynamic simulations in exoskeleton design and control.

Researchers have extensively investigated the intricate issue of musculoskeletal-exoskeletal multibody simulation, employing various methods such as inverse, forward, or a combination of both. Inverse dynamics methods, commonly employed in robotics and human biomechanics research, utilize known motions as inputs to the model to determine the coordinates (e.g., joint angles) and forces (e.g., muscle forces and joint torques) that are consistent with external measurements (e.g., trajectories from a motion capture system and ground reaction forces) [11–16]. These methods prove valuable in analyzing muscle activation and force changes in the upper and lower extremities during specific motions while wearing an exoskeleton [17–20]. However, inverse dynamics approaches rely on kinematic data to generate motion, failing to provide insights into the underlying neuromuscular control strategies responsible for movement patterns and limiting our understanding of movement biomechanics. Additionally, when applied to solving musculoskeletal models, inverse dynamics methods often assume time-independent human motions and employ static optimization (SO) techniques [21] to address muscle redundancy, overlooking factors such as force balance within the muscle-tendon unit and muscle-contraction dynamics. Moreover, reliance on measured foot-ground reaction forces for calculating muscle forces and joint torques imposes significant demands on measurement equipment like force plates[22], thus confining related studies to laboratory or simplified environments. It should be noted that the scarcity of simulation platforms capable of effectively adjusting foot-ground interaction impedes the comprehensive multibody dynamics simulations of musculoskeletal-exoskeletal systems across a variety of environments.

On the other hand, forward dynamics approaches offer an alternative by utilizing muscle feedback signals obtained from surface electromyography (EMG) [23–26] or neural controllers [27] as inputs to the model. These approaches employ an optimal control framework to predict the motion (e.g., joint angles) and forces (e.g., muscle forces, joint torques, and ground reaction forces) of coupled musculoskeletal-exoskeletal systems. Researchers have developed various dynamic optimization methods to minimize the metabolic cost of the human body in simulations and reduce the errors associated with kinematic data or constraints [28,29]. However, it is worth noting that numerical algorithms used in forward dynamics analysis (e.g., shooting and collocation methods) are often computationally less efficient compared to inverse dynamics. As a result, these methods often compute only one gait cycle to represent human motion. However, in real-life scenarios, human gait needs to adapt to environmental changes, and sacrificing the number of gait cycles for computational efficiency may not be reasonable for simulating the motion of a coupled human-exoskeleton system in outdoor environments. To tackle this challenge, the covariance matrix adaptive evolution strategy (CMA-ES) has emerged as a promising approach. CMA-ES utilizes its excellent parallel computing capability to evaluate multiple solutions simultaneously and accelerate the optimization process, making it suitable for solving complex and computationally demanding problems. In recent years, researchers have successfully applied the CMA-ES algorithm to solve complex nonlinear optimization problems of musculoskeletal forward dynamics models [30,31]. Although the abovementioned optimization-based forward dynamic frameworks have proven effective in analyzing the dynamics of human movement, they often overlook the investigation of the neural-muscular-skeletal control strategies that govern the generation of movement, thus limiting our understanding of the underlying biomechanical mechanisms of the system. To address this gap, it is imperative to develop a dynamic simulation framework that integrates a comprehensive neural-muscular-skeletal control loop. By doing so, this framework is expected to provide a more holistic approach to studying human movement by generating comprehensive outputs, including muscle activations, muscle forces, joint torques, ground reaction forces, and human-machine interactions. These outputs will facilitate a detailed analysis of the kinematics and biomechanics of the system, leading to a deeper understanding of human locomotion.

Furthermore, to improve the accuracy and computational efficiency of the models, researchers have developed hybrid frameworks that combine the merits of forward and inverse dynamics methods [32,33]. One approach is the force-dependent kinematics (FDK) method, implemented in AnyBody software, which integrates inverse dynamics and quasi-static analysis to calculate joint torques but does not account for the muscle-tendon equilibrium equations. Another widely used hybrid method is the computed muscle control (CMC) [34,35], which improves the accuracy of the static optimization (SO) model by incorporating a PID controller that feeds the forward dynamics model's error back to the SO model. Nonetheless, CMC sometimes exhibits poor convergence [36]. The forward-muscular inverse-skeletal (FMIS) framework is also an effective hybrid dynamic simulation method [33]. It generates muscle activation and muscle contraction dynamics in a forward manner while solving skeletal dynamics in an inverse manner. The FMIS framework [37] introduces torque-tracking errors into the objective function to relax dynamic equilibrium constraints, thereby enhancing motion and torque tracking while improving computational efficiency. However, the torque tracking mode in FMIS does not align with the motion mechanisms generated by human neuromusculoskeletal control. It is important to note that although hybrid methods combining forward and inverse dynamics offer advantages, the reliance on kinematic data remains a limitation for estimating the dynamics of the human-exoskeleton coupled system outside laboratory settings. Moreover, these approaches disrupt the transmission order in the human neuromusculoskeletal feedback control loop, hindering the exploration of human movement biomechanics.

The aforementioned dynamics methods have laid a foundation for simulating the dynamics of human-exoskeleton coupling systems. However, each method has inherent limitations and falls short of providing a comprehensive assessment of system performance and biomechanics in diverse environments and tasks. Currently, there is a lack of integrated and comprehensive dynamic model and simulation framework that consider the human neuromusculoskeletal loop, exoskeleton, human-machine coupling, and environmental factors. To address this gap, this study presents an advanced and integrated dynamic simulation framework based on forward dynamics. This framework combines the human neuromusculoskeletal system with the exoskeleton and incorporates adjustable human-machine coupling and foot-ground contact. It goes beyond providing information on joint kinematics and extends to analysis of biomechanics, including muscle activations, muscle forces, and joint torques, by integrating a complete neuromusculoskeletal feedback loop. Furthermore, the proposed simulation framework offers valuable insights into the interaction forces between the human and exoskeleton, which are crucial for analyzing the assistance provided by exoskeletons during human motions. Notably, this simulation framework does not rely on external measurements such as kinematic data and ground reaction forces as inputs, enabling comprehensive multibody dynamics simulations of human-exoskeleton coupled systems in a wide range of environments.

Specifically, the proposed simulation framework adopts a multi-degree-of-freedom multi-body system to model the coupled human-exoskeleton system. Comprehensively, it encompasses essential components including human lower limb muscle reflex mechanisms, muscle activation dynamics, muscle-tendon contraction dynamics, musculoskeletal dynamics, human body-exoskeleton interaction, and system-environment contact. The dynamic model is solved through an optimization-based process that introduces a multi-objective fitness function, which takes into account key gait criteria such as metabolic cost, walking stability, and joint hyperextension. The free muscle reflex parameters are then optimized using a modified CMA-

ES with reference to neurophysiology. Our framework, which incorporates the complete neuromusculoskeletal feedback loop, allows for a comprehensive assessment of biomechanical and kinematic signals, including muscle activations, muscle forces, and joint torques, etc. These quantities, which are often challenging to measure experimentally, plays a vital role in advancing our understanding of human biomechanics and evaluating the efficacy of exoskeletons in facilitating human locomotion. To validate the proposed simulation framework, a comparison is conducted between the simulation results and the experimental data obtained from a healthy subject wearing an exoskeleton while walking at different speeds and terrains. Although a quantitative assessment of the subject's behavior and biomechanics is neither possible nor reasonable due to challenges in precisely measuring physical parameters such as segment masses and moments of inertia, as well as the uniqueness of walking habits, the results indicate that the simulation framework effectively captures the qualitative aspects of the coupled system's kinematic and biomechanical signals. The results highlight the effectiveness and accuracy of our framework in capturing qualitative differences in muscle activity related to different functions, as well as the evolutionary patterns of muscle activity and kinematic signals across different walking speeds and terrains. With these merits, the developed dynamic simulation framework enables efficient and cost-effective performance testing of new or improved exoskeleton designs and control strategies. This eliminates the need for frequent prototyping of new exoskeletons and conducting expensive human experiments. Furthermore, by incorporating adjustable human-environment interactions within the simulation framework, this approach provides a convenient method for analyzing the improvements in exoskeleton-assisted human locomotion performance across diverse terrains.

## 2. Methods

### 2.1 Overview of the framework

**Fig. 1(A)** illustrates the working principle of the forward dynamics simulation framework for coupled neuromusculoskeletal-exoskeletal systems. The framework consists of two subsystems: the muscle dynamics subsystem, encompassing the muscle reflex mechanism, muscle activation dynamics, musculotendon contraction dynamics, and musculoskeletal kinematics, and the rigid-body dynamics subsystem, representing the multi-rigid-body dynamics of the human-exoskeleton system. The rigid-body dynamics subsystem also incorporates the interaction between the human and exoskeleton, as well as the reaction forces exerted by the ground on the coupled system. A key innovation of the simulation framework lies in the approach to determining internal signals, including muscle activation, muscle forces, and joint torques. Rather than relying on empirical pre-determination or direct optimization methods, these signals are derived from a physiological perspective. This is achieved by optimizing the muscle reflex parameters, taking into account the intricate interplay among muscle activities, human-exoskeleton interaction, and the dynamic interaction between the system and the environment. Such an approach ensures a more accurate representation of the physiological behavior within the simulation framework.

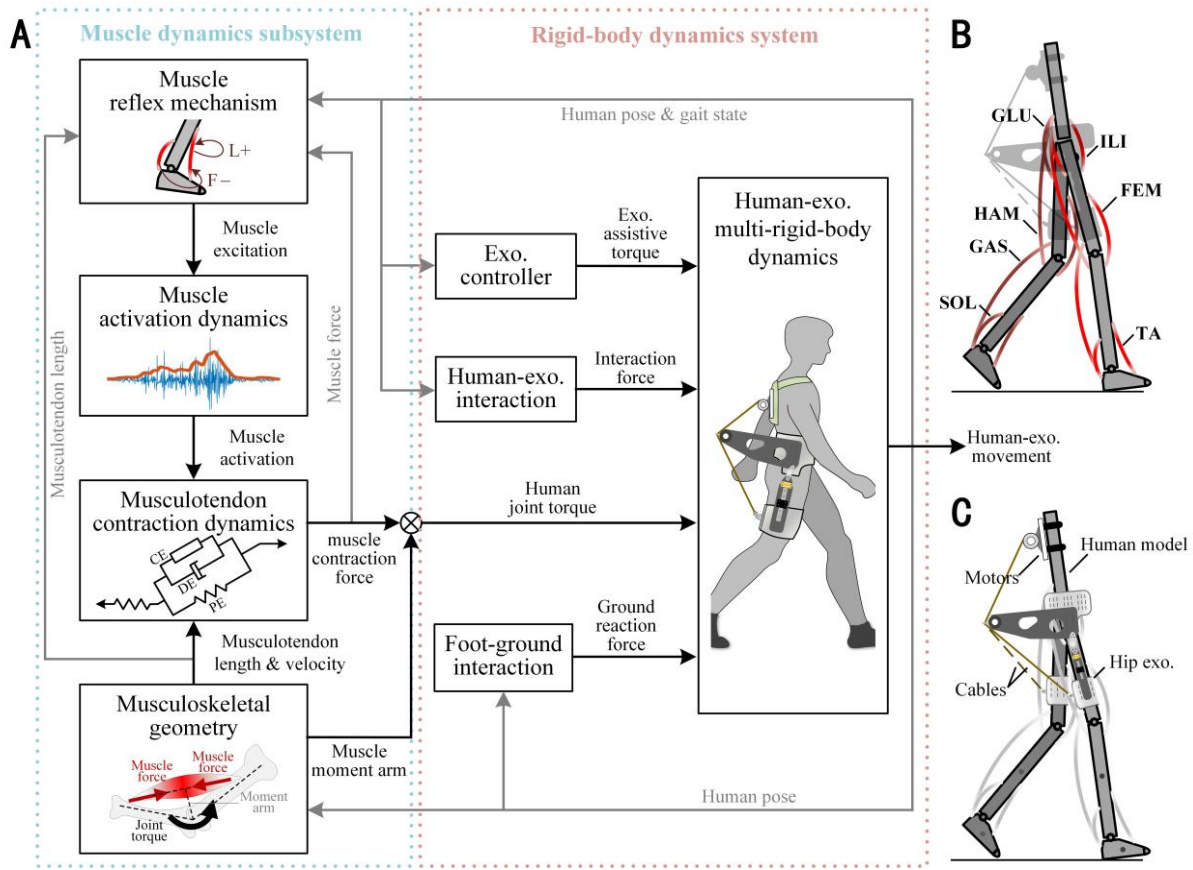

**Fig. 1.** Overview of the forward dynamics simulation framework for coupled neuromusculoskeletal-exoskeletal systems. (A) Flow diagram of the framework, demonstrating the working principle, where the grey arrows indicate the feedback. (B) Illustration of leg muscles, where the brown and red curves indicate the muscles of the left and right leg, respectively (GLU: gluteus maximus; ILI: iliopsoas; FEM: rectus femoris; HAM: hamstrings; GAS: gastrocnemius; SOL: soleus; TA: tibialis anterior). (C) Illustration of human-exoskeleton coupling, exemplified by a cable-driven hip exoskeleton.

Muscle excitation, referring to the efferent signal transmitted by the motor neurons of the central nervous system, is generated through the muscle reflex mechanism. This mechanism takes into account various kinematic and physiological factors such as human pose, gait phases, musculotendon lengths, and muscle forces. The muscle excitation is then transformed into muscle activation through the process of muscle activation dynamics. Simultaneously, utilizing the information on human-exoskeleton movement and muscle position, the length and velocity of the musculotendon, as well as the moment arm of the force relative to the rotation axis of the corresponding joint, can both be calculated through musculoskeletal kinematics.

Subsequently, the muscle activation, along with musculotendon length and velocity, drives the muscle to contract by employing the musculotendon contraction dynamics, which is simulated by the Hill-type muscle-tendon model [38].

By multiplying musculotendon contraction force with the moment arm of the muscle at each time instant, human joint torques can be derived. Furthermore, the ground reaction force and the interaction force between the human body and the exoskeleton can be customized based on user-defined models. Similarly, the assistive torque of the exoskeleton can be tailored according to the given exoskeleton control strategy and the current gait state. The ability to customize the human-exoskeleton interaction and foot-ground interaction is a notable advantage of the proposed simulation framework.

The calculated human joint torques, human-exoskeleton interaction forces, ground reaction forces, and exoskeleton assistive torques are then employed as inputs to the human-exoskeleton multi-rigid-body model. This integration facilitates the movement of the coupled human-exoskeleton model, enabling a comprehensive analysis of the dynamic behavior.

To capture the influence of the key lower limb muscles involved in human walking in the sagittal plane at the neuromusculoskeletal level [39], the framework incorporates seven muscles in each leg, as depicted in **Fig. 1(B)** for reference. These muscles are crucial for controlling various movements. Specifically, five uniarticular muscles are identified: the iliopsoas (ILI) controls hip flexion, the gluteus maximus (GLU) controls hip extension, the rectus femoris (FEM) controls knee extension, the soleus (SOL) controls plantarflexion, and the tibialis anterior (TA) controls dorsiflexion. Additionally, two biarticular muscles, namely the hamstrings (HAM) and gastrocnemius (GAS), are also included in the framework; the hamstrings play a role in both hip extension and knee flexion, while the gastrocnemius contributes to knee flexion and plantarflexion movements.

The presented forward dynamics simulation framework also offers the flexibility to customize the exoskeleton hardware. In this study, a cable-driven hip exoskeleton is utilized to exemplify the capabilities and performance of the proposed simulation framework. As depicted in **Fig. 1(C)**, the cable-driven assistive hip exoskeleton model features two portable motors positioned on the user's back. These motors generate assistive forces that are transmitted to the thigh brace via cables. By considering the geometry of the coupled human-exoskeleton model, these forces can be effectively converted into assistive torques.

## 2.2 Modeling of muscle reflex and activation

### *2.2.1 Muscle reflex model*

Physiologically, the activation of human skeletal muscles is governed by electrochemical signals known as muscle excitations $\sigma$, which are transmitted from the central nervous system to muscle fibers. The muscle reflex model provides a mathematical representation of how the states of the legs and muscles influence muscle excitation during voluntary movements. By incorporating the principles of legged mechanics and refining the segmentation of the gait, we develop an enhanced model of voluntary muscle reflexes, building upon the studies conducted by Geyer and Herr [27] and Geijtenbeek et al. [40]. The improved model, illustrated in **Fig. 2**, provides a more detailed and comprehensive representation of the muscle reflex during locomotion. In this section, we will delve into the muscle reflex mechanism, providing a detailed explanation, and outlining the modifications we have introduced.

In our study, we divide a single gait cycle into five distinct phases: three stance phases (early stance, late stance, and liftoff) and two swing phases (early swing and landing). Conventionally, a complete gait cycle is defined as the duration between two consecutive heel strikes of one foot [39]. Transitions between these gait phases are determined by a comprehensive assessment of the ground force and foot position of both legs, building upon the approach proposed by Geijtenbeek et al. [40].

To facilitate subsequent descriptions, we will first provide the definitions of the following symbols: $\sigma$ denotes the muscle excitation, $K_L$, $K_F$ are positive proportional coefficients of muscle length and muscle force, $\tilde{l}$, $\tilde{l}_0$, and $\tilde{F}$ represent the normalized muscle length, normalized reference muscle length, and normalized muscle force, respectively. In the reflex model, the normalization is usually performed with respect to the optimal muscle length and optimal muscle force, i.e., $\tilde{l} = l/l_{\text{opt}}$, $\tilde{F} = F/F_{\text{opt}}$. The operator $\{\}_+$ is defined as follows: it takes a positive value when the variable is positive, and 0 when the variable is non-positive. In the subsequent discussion, the superscript on the right indicate the specific muscle of interest.

The reflex mechanism of the TA muscle during a gait cycle is defined by

$$\sigma^{\text{TA}} = \left\{ K_L^{\text{TA}} \left( \tilde{l}^{\text{TA}} - \tilde{l}_0^{\text{TA}} \right) - K_F^{\text{TA}} \tilde{F}^{\text{SOL}} \right\}_+ . \tag{1}$$

In Eq. (1), the first term $K_L^{\text{TA}} \left( \tilde{l}^{\text{TA}} - \tilde{l}_0^{\text{TA}} \right)$ represents a positive muscle length feedback (L+) from TA to $\sigma^{\text{TA}}$, indicating that an increase in the length of TA can enhance its excitation; while the second term $-K_F^{\text{TA}} \tilde{F}^{\text{SOL}}$ denotes a negative muscle force feedback (F–), indicating that the contraction of SOL can inhibit the excitation of TA and providing a mechanism through which SOL activity can modulate the excitation of TA. The L+ mechanism of TA ensures foot clearance during the swing phases, and the F− mechanism demonstrates the antagonistic relationship between TA and SOL.

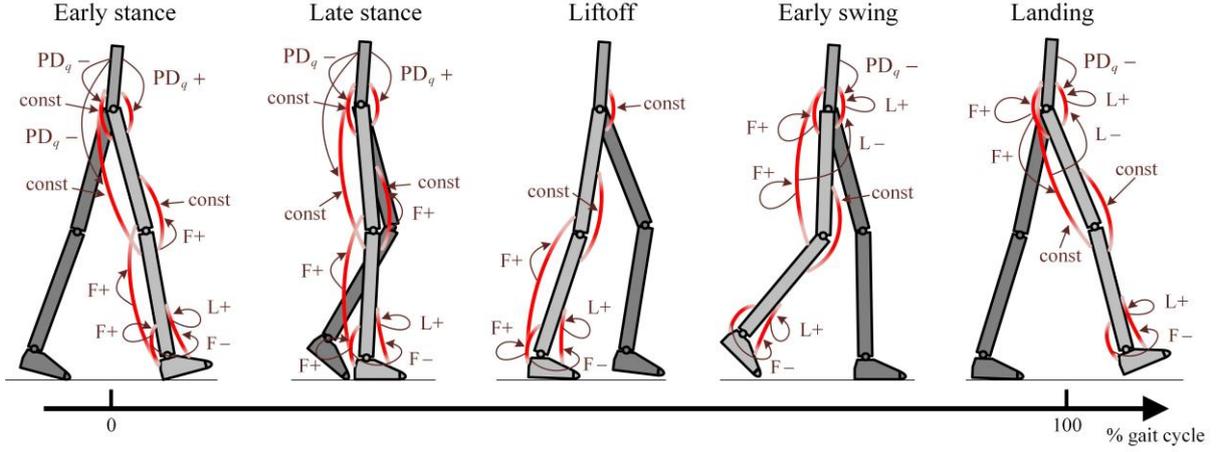

**Fig. 2.** Reflex mechanism of leg muscles during different gait phases, where the leg muscles along with the corresponding feedbacks and constant stimulations that influence muscle excitations are indicated.

The reflex mechanism of the SOL muscle is given by

$$\sigma^{\text{SOL}} = \begin{cases} \{K_F^{\text{SOL}} \tilde{F}^{\text{SOL}}\}_+, & \text{early stance / late stance / liftoff,} \\ 0, & \text{early swing / landing,} \end{cases} \quad (2)$$

and the mechanism of the GAS muscle exhibits similarities to the SOL mechanism

$$\sigma^{\text{GAS}} = \begin{cases} \{K_F^{\text{GAS}} \tilde{F}^{\text{GAS}}\}_+, & \text{early stance / late stance / liftoff,} \\ 0, & \text{early swing / landing.} \end{cases} \quad (3)$$

Eq. (2) and (3) indicate that both SOL and GAS are activated by positive force feedback (F+) from their own contraction during the stance phases. This activation makes SOL and GAS the main contributors to foot propulsion. However, during the swing phases of the corresponding leg, these muscles remain inactive.

The reflex mechanism of the FEM muscle is defined by

$$\sigma^{\text{FEM}} = \begin{cases} \{C_1^{\text{FEM}} + \xi K_{F_1}^{\text{FEM}} \tilde{F}^{\text{FEM}}\}_+, & \text{early stance,} \\ \{C_1^{\text{FEM}} + \xi K_{F_2}^{\text{FEM}} \tilde{F}^{\text{FEM}}\}_+, & \text{late stance,} \\ \{C_2^{\text{FEM}}\}_+, & \text{liftoff / early swing / landing,} \end{cases} \quad (4)$$

where $C_1^{\text{FEM}}$ and $C_2^{\text{FEM}}$ are constant stimulations to the muscle (const); $K_{F_1}^{\text{FEM}} \tilde{F}^{\text{FEM}}$ and $K_{F_2}^{\text{FEM}} \tilde{F}^{\text{FEM}}$ represent positive force feedbacks (F+) that induce eccentric contraction of FEM, thereby helping to maintain knee stiffness during the early stance and late stance phases. The symbol $\xi$, which is equal to 0 or 1, represents a conditional switch, defined by

$$\xi = \begin{cases} 1, & q^{\text{knee}} < \hat{q}^{\text{knee}} \text{ or } \dot{q}^{\text{knee}} \leq 0, \\ 0, & \text{else,} \end{cases} \quad (5)$$

where $q^{\text{knee}}$, $\dot{q}^{\text{knee}}$, and $\hat{q}^{\text{knee}}$ are the angle, angular velocity, and angle threshold of the knee joint, respectively, with the direction of knee flexion as positive. When the knee is flexing above the threshold, the conditional switch $\xi$ is set to 0, deactivating the F+ feedback and suppressing the muscle force of FEM. This mechanism ensures prompt inhibition of FEM, preventing hyperextension of the knee joint. After the late stance phase, the muscle is stimulated at a constant value $C_2^{\text{FEM}}$ (const).

The reflex mechanism of the HAM muscle is characterized by

$$\sigma^{\text{HAM}} = \begin{cases} \left\{ C^{\text{HAM}} - K_q^{\text{HAM}} \left( q^{\text{UB}} - q_0^{\text{UB}} \right) - K_{\dot{q}}^{\text{HAM}} \dot{q}^{\text{UB}} \right\}_+, & \text{early stance / late stance,} \\ 0, & \text{liftoff / early swing,} \\ \left\{ C^{\text{HAM}} + K_F^{\text{HAM}} \tilde{F}^{\text{GLU}} \right\}_+, & \text{landing,} \end{cases} \quad (6)$$

where $q^{\text{UB}}$, $\dot{q}^{\text{UB}}$, and $q_0^{\text{UB}}$ are the angle, angular velocity, and reference angle of the upper body, respectively. $-K_q^{\text{HAM}} \left( q^{\text{UB}} - q_0^{\text{UB}} \right) - K_{\dot{q}}^{\text{HAM}} \dot{q}^{\text{UB}}$ represents a negative proportional-derivative (PD) feedback concerning the tilt of the upper body ($\text{PD}_q -$) during the early stance and late stance phases, which plays a crucial role in ensuring the stability of the upper body. Furthermore, the contraction of GLU can activate HAM during the landing phase through positive force feedback $K_F^{\text{HAM}} \tilde{F}^{\text{GLU}}$ (F+), indicating a synergy relationship between the two muscles. HAM is additionally stimulated at a constant value $C^{\text{HAM}}$ during the early stance, late stance, and landing phases.

The reflex mechanism of the GLU muscle is defined by

$$\sigma^{\text{GLU}} = \begin{cases} \left\{ C^{\text{GLU}} - K_q^{\text{GLU}} \left( q^{\text{UB}} - q_0^{\text{UB}} \right) - K_{\dot{q}}^{\text{GLU}} \dot{q}^{\text{UB}} \right\}_+, & \text{early stance / late stance,} \\ 0, & \text{liftoff,} \\ \left\{ K_F^{\text{GLU}} \tilde{F}^{\text{GLU}} \right\}_+, & \text{early swing / landing.} \end{cases} \quad (7)$$

GLU is excited by a constant stimulation $C^{\text{GLU}}$ (const), along with negative PD feedback $-K_q^{\text{GLU}} \left( q^{\text{UB}} - q_0^{\text{UB}} \right) - K_{\dot{q}}^{\text{GLU}} \dot{q}^{\text{UB}}$ ($\text{PD}_q -$) during the early stance and late stance phases. The PD feedback is responsible for maintaining the global orientation of the upper body. During the swing phases, a positive force feedback $K_F^{\text{GLU}} \tilde{F}^{\text{GLU}}$ (F+) from itself is applied to GLU, decelerating the thigh.

The reflex mechanism of the ILI muscle is given by

$$\sigma^{\mathrm{ILI}} = \begin{cases} \left\{ C_1^{\mathrm{ILI}} + K_{q_1}^{\mathrm{ILI}} \left( q^{\mathrm{UB}} - q_0^{\mathrm{UB}} \right) + K_{\dot{q}_1}^{\mathrm{ILI}} \dot{q}^{\mathrm{UB}} \right\}_+, & \text{early stance / late stance,} \\ C_2^{\mathrm{ILI}}, & \text{liftoff,} \\ \left\{ \begin{array}{l} K_{L_1}^{\mathrm{ILI}} \left( \tilde{l}^{\mathrm{ILI}} - \tilde{l}_0^{\mathrm{ILI}} \right) - K_{q_2}^{\mathrm{ILI}} \left( q^{\mathrm{UB}} - q_1^{\mathrm{UB}} \right) \\ - K_{\dot{q}_2}^{\mathrm{ILI}} \dot{q}^{\mathrm{UB}} - K_{L_2}^{\mathrm{ILI}} \left( \tilde{l}^{\mathrm{HAM}} - \tilde{l}_0^{\mathrm{HAM}} \right) \end{array} \right\}_+, & \text{early swing / landing.} \end{cases} \quad (8)$$

In the early stance and late stance phases, the excitation of ILI is regulated by a positive PD feedback $K_{q_1}^{\mathrm{ILI}} \left( q^{\mathrm{UB}} - q_0^{\mathrm{UB}} \right) + K_{\dot{q}_1}^{\mathrm{ILI}} \dot{q}^{\mathrm{UB}}$ ($\mathrm{PD}_q +$), along with constant stimulation $C_1^{\mathrm{ILI}}$ (const). Similar to HAM and GLU, the PD feedback for ILI also plays a role in maintaining the upper body orientation. During liftoff, a constant stimulation $C_2^{\mathrm{ILI}}$ (const) is applied to enhance the hip flexion torque that elevates the leg forward. During the swing phases, ILI is activated by a combined mechanism, involving negative PD feedback $-K_{q_2}^{\mathrm{ILI}} \left( q^{\mathrm{UB}} - q_1^{\mathrm{UB}} \right) - K_{\dot{q}_2}^{\mathrm{ILI}} \dot{q}^{\mathrm{UB}}$ ($\mathrm{PD}_q -$), positive length feedback from itself $K_{L_1}^{\mathrm{ILI}} \left( \tilde{l}^{\mathrm{ILI}} - \tilde{l}_0^{\mathrm{ILI}} \right)$ (L+), and negative length feedback from HAM $-K_{L_2}^{\mathrm{ILI}} \left( \tilde{l}^{\mathrm{HAM}} - \tilde{l}_0^{\mathrm{HAM}} \right)$ (L–).

By implementing a refined segmentation of gait phases, our model effectively captures the distinct behavior of key muscles during specific phases of the gait cycle. Specifically, we distinguish the behavior of the FEM muscle in early stance and late stance (Eq. (4)), the HAM muscle in early swing and landing (Eq. (6)), and the GLU muscle in liftoff (Eq. (7)). This refinement, in comparison to Geijtenbeek et al. [40], enhances the representation accuracy of muscle reflexes and improves the fidelity of muscle excitation in dynamic simulations.

The muscle reflex models developed above provide a mathematical description and interpretation of the reflex mechanism of each muscle. There are 29 parameters to be determined by optimization to capture the characteristics of human motion, and they constitute the reflex parameter vector $\mathbf{w} \in \mathbb{R}^{29}$. In detail, $\mathbf{w}$ is comprised of proportional coefficients ($K_L^{\mathrm{TA}}, K_F^{\mathrm{TA}}, K_F^{\mathrm{SOL}}, K_F^{\mathrm{GAS}}, K_{F_1}^{\mathrm{FEM}}, K_{F_2}^{\mathrm{FEM}}, K_q^{\mathrm{HAM}}, K_{\dot{q}}^{\mathrm{HAM}}, K_F^{\mathrm{HAM}}, K_q^{\mathrm{GLU}}, K_{\dot{q}}^{\mathrm{GLU}}, K_F^{\mathrm{GLU}}, K_{q_1}^{\mathrm{ILI}}, K_{\dot{q}_1}^{\mathrm{ILI}}, K_{L_1}^{\mathrm{ILI}}, K_{q_2}^{\mathrm{ILI}}, K_{\dot{q}_2}^{\mathrm{ILI}}, K_{L_2}^{\mathrm{ILI}}$), normalized reference muscle lengths ($\tilde{l}_0^{\mathrm{TA}}, \tilde{l}_0^{\mathrm{ILI}}, \tilde{l}_0^{\mathrm{HAM}}$), joint angle thresholds ($q_0^{\mathrm{UB}}, \hat{q}_0^{\mathrm{knee}}$), and constant stimulations ($C_1^{\mathrm{FEM}}, C_2^{\mathrm{FEM}}, C^{\mathrm{HAM}}, C^{\mathrm{GLU}}, C_1^{\mathrm{ILI}}, C_2^{\mathrm{ILI}}$). The detailed optimization procedures will be introduced in Section 3.1.

*2.2.2 Muscle activation model*

The level of muscle activation, denoted as $a$, depicts the concentration of bound Ca2+-ions in the muscle sarcoplasm relative to its physiological maximum, and typically ranges from $a_{\min}$ to 1 [41]. However, it is important to note that the activation and deactivation of the muscle do not occur instantaneously following muscle excitation. The time delay between muscle excitation $\sigma$ and muscle activation $a$ can be described by a simplified first-order dynamic

model [42]:

$$\dot{a} = \frac{\sigma - a}{\tau(a,\sigma)}, \tag{9}$$

$$\tau(a,\sigma) = \begin{cases} \tau_{\text{act}}(0.5 + 1.5a), & \sigma > a, \\ \tau_{\text{deact}}/(0.5 + 1.5a), & \sigma \leq a, \end{cases} \tag{10}$$

where $\tau_{\text{act}}$ and $\tau_{\text{deact}}$ denote the activation and deactivation delay constant, respectively. These delays account for the time required for the propagation of the nerve action potential, chemical transmission at the neuromuscular junction, and subsequent steps involved in muscle activation. It is worth mentioning that although more detailed activation models have been proposed in [43,44], this simplified model offers computational efficiency that is suitable for addressing complex challenges, such as gait simulation in a coupled neuromusculoskeletal-exoskeletal dynamic system.

## 2.3 Modeling of musculotendon contraction dynamics

To accurately depict the process of converting muscle excitation into muscle force, several factors such as muscle activation, muscle fiber length and velocity, tendon length and velocity, and others need to be considered. In this study, we employ the modified Hill-type model developed by Millard et al [45] (refer to **Fig. 3(A)**) to describe musculotendon contraction dynamics. For the sake of convenience and comprehensiveness, we briefly summarize this model in the following.

The model incorporates four elements to represent a muscle:
- A *contractile element* (CE) responsible for generating muscle fiber contraction in response to muscle activation.
- A *damper element* (DE) connected in parallel with CE to capture the damping characteristics of the muscle.
- A *parallel elastic element* (PE) that generates a passive elastic force in the muscle.
- A *tendon* (T) that captures the elastic properties of the tendon connected in series with the muscle fiber.

The active force produced by CE depends on four key factors: the optimal muscle force $F_{\text{opt}}$, the muscle activation $a$, muscle length $l^{\text{M}}$, and muscle contraction velocity $v^{\text{M}}$, governed by:

$$\tilde{F}^{\text{CE}} = F^{\text{CE}}/F_{\text{opt}} = a\mathbf{f}^{\text{L}}(\tilde{l}^{\text{M}})\mathbf{f}^{\text{V}}(\tilde{v}^{\text{M}}), \tag{11}$$

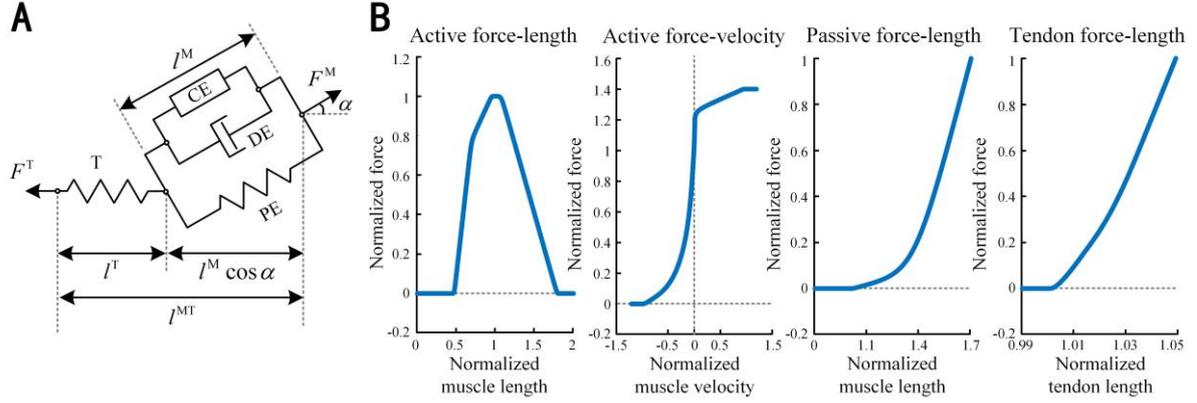

**Fig. 3.** The modified Hill-type muscle model describing the musculotendon contraction dynamics. (A) The four elements incorporated in the muscle model. The figure also shows the length of individual elements and the pennation angle. (B) Four characteristic curves that describe the muscle behavior, from the left to the right showing the relationships of active force-length, active force-velocity, passive force-length, and tendon force-length.

where $\tilde{l}^M$ is the muscle length normalized by its optimal value, i.e., $\tilde{l}^M = l^M/l_{opt}$; and $\tilde{v}^M$ is the derivative of $\tilde{l}^M$ with respect to time, i.e., $\tilde{v}^M = \dot{\tilde{l}}^M$; $\mathbf{f}^L(\tilde{l}^M)$ depicts the relationship between the active force and muscle length, and $\mathbf{f}^V(\tilde{v}^M)$ represents the relationship between the active force and the muscle contraction velocity (see **Fig. 3(B)**). Typically, the muscle produces a greater amount of isometric force as it approaches its optimal length. The force peaks at $F_{opt}$ when the muscle length is $l^M_{opt}$.

DE generates resistive force proportional to the contraction or relaxation velocity of the muscle:

$$\tilde{F}^{DE} = F^{DE}/F_{opt} = \beta\,\tilde{v}^M, \tag{12}$$

where $\beta$ is a proportionality coefficient. In the past, this element has seldom been taken into account when estimating the force production of the Hill-type muscle. However, considering that the human striated muscle contains up to 79% water [45,46], it is reasonable to incorporate the damping effect.

The passive force produced by PE, and the tendon force produced by tendon (T), are respectively modeled as non-linear springs based on their length:

$$\tilde{F}^{PE} = F^{PE}/F_{opt} = \mathbf{f}^{PE}(\tilde{l}^M), \tag{13}$$

$$\tilde{F}^T = F^T/F_{opt} = \mathbf{f}^T(\tilde{l}^T), \tag{14}$$

where $\mathbf{f}^{PE}(\tilde{l}^M)$ depicts the relationship between the passive muscle force and muscle length, and $\mathbf{f}^T(\tilde{l}^T)$ represents the relationship between the tendon force and tendon length (see **Fig. 3(B)**). $C_2$-continuous Quintic Bézier splines have been utilized to represent the four characteristic curves [47]. In Eq. (14), $\tilde{l}^{\text{T}}$ is the tendon length normalized by its stack length $l_{\text{S}}^{\text{T}}$ (slack length refers to the length of the tendon when no force is applied to it), i.e., $\tilde{l}^{\text{T}} = l^{\text{T}}/l_{\text{S}}^{\text{T}}$. The total length of the musculotendon unit is $l^{\text{MT}}$ is determined based on the posture of human movement and the attachment position of the musculotendon to the bone, it can be expressed by

$$l^{\text{MT}} = l^{\text{T}} + l^{\text{M}} \cos\alpha, \tag{15}$$

where $\alpha$ depicts the pennation angle, i.e. the instantaneous angle between the muscle fiber and the tendon. As the muscle shortens, the distance $l^{\text{M}} \sin\alpha$ remains constant, and as a result, $\alpha$ increases. Therefore, $\alpha$ can be obtained by considering the following equation:

$$l^{\text{M}} \sin\alpha = l_{\text{opt}} \sin\alpha_{\text{opt}}, \tag{16}$$

where $\alpha_{\text{opt}}$ is the pennation angle when the muscle length stays at the optimal value.

The modified Hill-type muscle model incorporates CE, DE, and PE in parallel. Therefore, the total force produced by a muscle, denoted as $F^{\text{M}}$, can be calculated via:

$$\begin{aligned} F^{\text{M}} &= F_{\text{opt}} \left( \tilde{F}^{\text{CE}} + \tilde{F}^{\text{DE}} + \tilde{F}^{\text{PE}} \right) \\ &= F_{\text{opt}} \left( a\mathbf{f}^{\text{L}}\left(\tilde{l}^{\text{M}}\right)\mathbf{f}^{\text{V}}\left(\tilde{v}^{\text{M}}\right) + \beta \tilde{v}^{\text{M}} + \mathbf{f}^{\text{PE}}\left(\tilde{l}^{\text{M}}\right) \right). \end{aligned} \tag{17}$$

In Eq. (17), $F_{\text{opt}}$ and $\beta$ are fixed physiological parameters specific to each muscle and are predetermined, $a$ is the muscle activation, which is determined through the muscle activation dynamics explained in subsection 2.2.1, and $\tilde{l}^{\text{M}}$ can be derived based on the initial muscle length and the muscle contraction velocity $\tilde{v}^{\text{M}}$. Hence, the muscle force can be solved if $\tilde{v}^{\text{M}}$ is obtained. To simplify the analysis, we assume the muscle mass is negligible, which implies that the difference between the tendon force and the muscle force along the tendon is zero, i.e.,

$$F_{\text{opt}} \left( a\mathbf{f}^{\text{L}}\left(\tilde{l}^{\text{M}}\right)\mathbf{f}^{\text{V}}\left(\tilde{v}^{\text{M}}\right) + \beta \tilde{v}^{\text{M}} + \mathbf{f}^{\text{PE}}\left(\tilde{l}^{\text{M}}\right) \right) \cos\alpha - F_{\text{opt}} \mathbf{f}^{\text{T}}\left(\tilde{l}^{\text{T}}\right) = 0. \tag{18}$$

Hence, the muscle contraction velocity yields

$$\tilde{v}^{\text{M}} = \mathbf{f}_{\text{inv}}^{\text{V}} \left( \frac{\mathbf{f}^{\text{T}}\left(\tilde{l}^{\text{T}}\right)/\cos\alpha - \beta \tilde{v}^{\text{M}} - \mathbf{f}^{\text{PE}}\left(\tilde{l}^{\text{M}}\right)}{a\mathbf{f}^{\text{L}}\left(\tilde{l}^{\text{M}}\right)} \right), \tag{19}$$

where $\mathbf{f}_{\text{inv}}^{\text{V}}$ is the inverse function of $\mathbf{f}^{\text{V}}$. In S1 of the **Supplementary Material**, detailed procedures for solving the muscle contraction velocity $\tilde{v}^{\text{M}}$ are provided.

## 2.4 Musculoskeletal modeling

### 2.4.1 Muscle moment derivation

In a forward-dynamic gait simulation, the motion of the human skeletal model is controlled by torques produced by multiple muscles spanning the joints [36]. The magnitude of torque generated at a joint is influenced by several factors, including the joint angle, the contraction level and the number of involved muscles, the attachment positions of the muscles, and their paths of action [48]. In this subsection, we focus on examining the influence of musculoskeletal geometry on muscle arms, which is crucial for accurately converting muscle forces into moments.

We utilized the musculoskeletal geometry provided by the SCONE H0914M model [49], which allows us to calculate the muscle arms around each joint. As shown in **Fig. 4**, two types of skeletal muscles are considered: the monoarticular muscle, which spans a single joint (e.g., SOL), and the biarticular muscle, which spans two joints (e.g., HAM). **Table 1** lists the variables employed in the calculation of muscle moments, where $\{A\}$, $\{B\}$ and $\{C\}$ represent the coordinates of the center of mass of bones A, B, and C, respectively; $\{W\}$ represents the world coordinate system, while $\{J\}$ represents the coordinate of the joint, which is transformed into $\{J'\}$ after rotating by an angle $\theta_J$.

### *Monoarticular muscle*

For the monoarticular muscle, the position vector of the muscle attachment site at bone A with respect to $\{J\}$ can be expressed as

$$\begin{bmatrix} {}^J\mathbf{p}_1 \\ 1 \end{bmatrix} = {}^J_A\mathbf{T} \begin{bmatrix} {}^A\mathbf{p}_1 \\ 1 \end{bmatrix}. \tag{20}$$

Similarly, the position vector of the muscle attachment site at bone B with respect to $\{J\}$ is

$$\begin{bmatrix} {}^J\mathbf{p}_2 \\ 1 \end{bmatrix} = \mathbf{R}(\theta_J) {}^{J'}_B\mathbf{T} \begin{bmatrix} {}^B\mathbf{p}_2 \\ 1 \end{bmatrix}, \tag{21}$$

where $\mathbf{R}(\theta_J)$ is the rotation matrix of $\{J'\}$ with respect to $\{J\}$:

$$\mathbf{R}(\theta_J) = \begin{bmatrix} \cos\theta_J & -\sin\theta_J & 0 \\ \sin\theta_J & \cos\theta_J & 0 \\ 0 & 0 & 1 \end{bmatrix}. \tag{22}$$

The muscle moment arm, which is the distance from the joint to the muscle (treated as a straight line), can be calculated via

$$d_{mus} = \frac{\left| {}^J\mathbf{p}_1 \times {}^J\mathbf{p}_2 \right|}{\left| {}^J\mathbf{p}_1 - {}^J\mathbf{p}_2 \right|}. \tag{23}$$

Note that once the type of muscle is determined, all variables of the monarticular muscle listed in **Table 1**, except for the joint angle $\theta_J$, remain fixed. Hence, the muscle moment arm is solely dependent on the joint angle $\theta_J$ and the type of muscle.

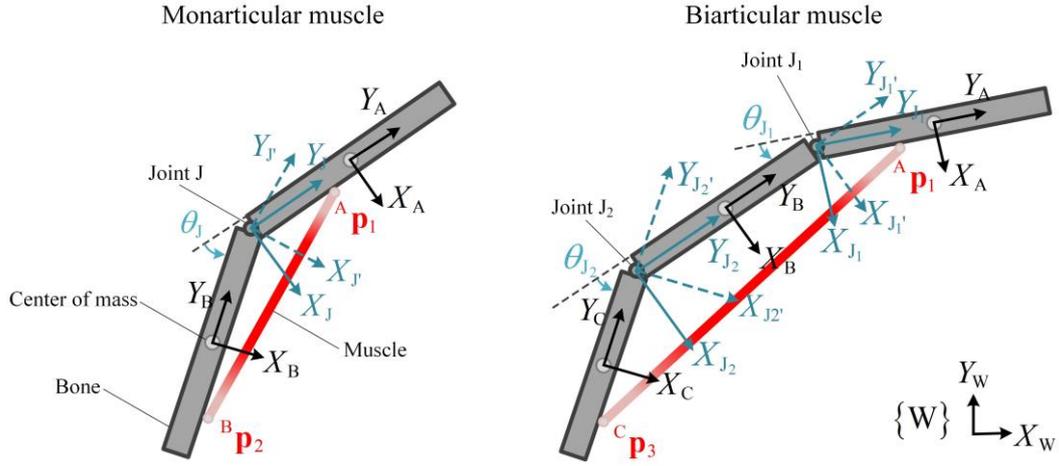

**Fig. 4.** Illustration of two types of musculoskeletal geometry.

Table 1 Variables involved in musculoskeletal geometry.

| Monarticular muscle | | Biarticular muscle | |
| --- | --- | --- | --- |
| Symbols | Descriptions | Symbols | Descriptions |
| $^A\mathbf{p}_1$ | 2×1 position vector of the muscle attachment site at bone A with respect to $\{A\}$ | $^A\mathbf{p}_1$ | 2×1 position vector of the muscle attachment site at bone A with respect to $\{A\}$ |
| $^J_A\mathbf{T}$ | 3×3 transformation matrix of $\{A\}$ with respect to $\{J\}$ | $^{J_1}_A\mathbf{T}$ | 3×3 transformation matrix of $\{A\}$ with respect to $\{J_1\}$ |
| $\theta_J$ | The angle of the joint, taking the counter-clockwise rotation as reference | $\theta_{J_1}$ | The angle of joint $J_1$, taking the counter-clockwise rotation as reference |
| $^{J'}_B\mathbf{T}$ | 3×3 transformation matrix of $\{B\}$ with respect to $\{J'\}$ | $^{J_1'}_B\mathbf{T}$ | 3×3 transformation matrix of $\{B\}$ with respect to $\{J_1'\}$ |
| $^B\mathbf{p}_2$ | 2×1 position vector of the muscle attachment site at bone B with respect to $\{B\}$ | $^B_{J_2}\mathbf{T}$ | 3×3 transformation matrix of $\{J_2\}$ with respect to $\{B\}$ |
| | | $\theta_{J_2}$ | The angle of joint $J_2$, taking the counter-clockwise rotation as reference |
| | | $^{J_2'}_C\mathbf{T}$ | 3×3 transformation matrix of $\{C\}$ with respect to $\{J_2'\}$ |
| | | $^C\mathbf{p}_3$ | 2×1 position vector of the muscle attachment site at bone C with respect to $\{C\}$ |

*Biarticular muscle*

Unlike monoarticular muscles, which cross over one joint, biarticular muscles spinning two joints present a more complex calculation for determining the muscle arm. The position vector of the muscle attachment site at bone A with respect to $\{J_1\}$ is

$$\begin{bmatrix} ^{J_1}\mathbf{p}_1 \\ 1 \end{bmatrix} = {}^{J_1}_A\mathbf{T} \begin{bmatrix} ^A\mathbf{p}_1 \\ 1 \end{bmatrix}, \tag{24}$$

and the position vector of the muscle attachment site at bone C with respect to $\{J_1\}$ is

$$\begin{bmatrix} {}^{J_1}\mathbf{p}_3 \\ 1 \end{bmatrix} = \mathbf{R}(\theta_{J_1}) \, {}^{J_1'}_{B}\mathbf{T} \, {}^{B}_{J_2}\mathbf{T} \cdot \mathbf{R}(\theta_{J_2}) \, {}^{J_2'}_{C}\mathbf{T} \begin{bmatrix} {}^{C}\mathbf{p}_3 \\ 1 \end{bmatrix}, \qquad (25)$$

where the rotation matrices $\mathbf{R}(\theta_{J_1})$ and $\mathbf{R}(\theta_{J_2})$ yield

$$\mathbf{R}(\theta_{J_1}) = \begin{bmatrix} \cos\theta_{J_1} & -\sin\theta_{J_1} & 0 \\ \sin\theta_{J_1} & \cos\theta_{J_1} & 0 \\ 0 & 0 & 1 \end{bmatrix}, \; \mathbf{R}(\theta_{J_2}) = \begin{bmatrix} \cos\theta_{J_2} & -\sin\theta_{J_2} & 0 \\ \sin\theta_{J_2} & \cos\theta_{J_2} & 0 \\ 0 & 0 & 1 \end{bmatrix}. \qquad (26)$$

The moment arms of the biarticular muscle around joints $J_1$ and $J_2$ can be calculated using a similar approach as in Eq. (23).

With the calculated muscle moment arm, the torque generated by muscle $i$ around the joint can be derived

$$\tau_i = F_i^M \cos\alpha \cdot d_{\text{mus}}, \qquad (27)$$

where $F_i^M \cos\alpha$ represents the component of force produced by muscle $i$ in the direction of the tendon. By summing up the torques produced by all related muscles, the joint torque can be calculated.

### *2.4.2 Multi-rigid-body dynamics of coupled human-exoskeleton systems*

Based on the modeling of muscle reflexes and activation, muscle-tendon contraction dynamics, and musculoskeletal models, we develop a multi-rigid-body dynamic model of the coupled human-exoskeleton system in this subsection. Specifically, we illustrate the modeling process with a human body wearing a powered cable-driven hip exoskeleton, which could provide assistive force during hip extension. **Fig. 5(A)** depicts the coupled human-exoskeleton system, in which the exoskeleton is coupled with the human body through waist and thigh braces. On each side of the body, a motor assembly is mounted on the exterior of the backpack, driving a cable that extends downward and connects to the thigh brace. A cable attachment, located on the waist strut, allows the cable to pass through it and functions as a fixed pulley as well as a motion-limiting mechanism [50]. The waist and thigh struts are bolted to the waist brace and the thigh brace, respectively, and are interconnected by a pin joint assembly. For the sake of modeling and analysis, the following assumptions are made:

- The relative displacements between the waist brace and the human trunk as well as the backpack and the human trunk are neglected because their interaction forces are relatively weak and do not significantly affect the response of the dynamical system. Therefore, the waist brace, the backpack, and the human trunk can be considered as a single rigid body.

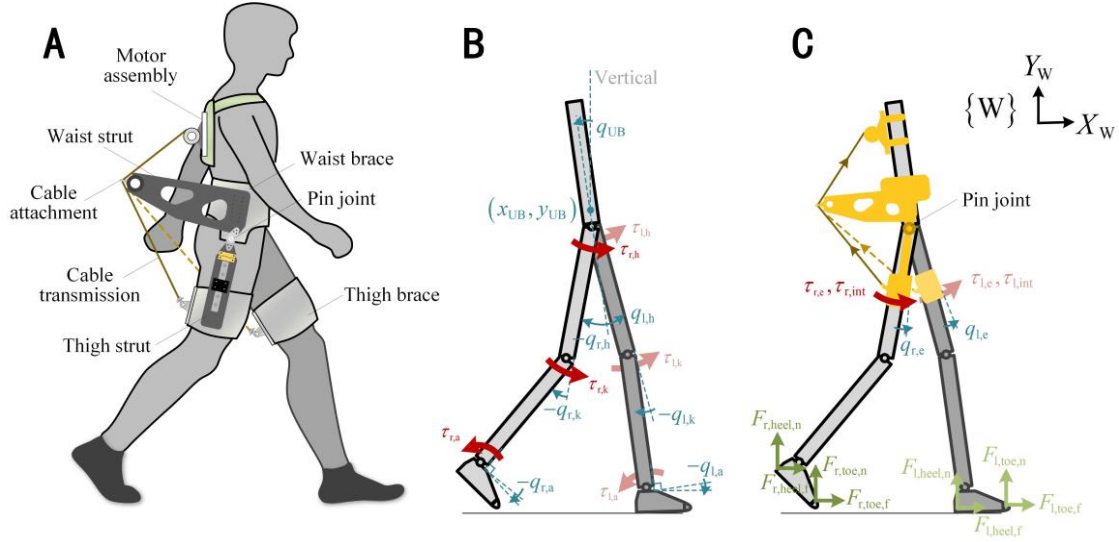

**Fig. 5**. Illustration of the coupled human-exoskeleton system. (A) A human wearing the powered cable-driven hip exoskeleton. (B) Generalized coordinates and joint torques of the human. (C) Generalized coordinates of the exoskeleton, interactive torques between the human and the exoskeleton, and the ground reaction forces.

- The motion, forces, and torques in the coronal and transverse planes are neglected due to their small range of motion [51].
- The axis of rotation of the pin joint in the exoskeleton and the axis of the human hip joint are ideally co-linear, thus preventing any joint misalignment issues in the model.

Based on the above assumptions, a dynamic model of the coupled human-exoskeleton system is developed **(Fig. 5(B) and (C))**, which consists of 9 rigid segments and is described by 11 DOFs. The joint torques are denoted as $\tau_{i_1,i_2}$, where the subscript $i_1 \in \{r, l\}$ represents right (r) and left (l), and $i_2 \in \{h, k, a\}$ represents hip (h), knee (k), and ankle (a). The exoskeleton produces assistive torques $\tau_{i_1,e}$ for the wearer, creating interaction torques $\tau_{i_1,\text{int}}$ with the thighs at the contact interface (the detailed form will be provided later). The angles between adjacent two rigid elements are denoted by $q_{i_1,i_2}$, and the angle between the upper body and the vertical plane is denoted by $q_{\text{UB}}$. Exceptionally, $q_{i_1,a}$ is defined as the angle between the foot and the line perpendicular to the shank, which aligns with the general definition of ankle plantarflexion/dorsiflexion [46]. Following the convention, positive torques and angles are defined as those that produce counter-clockwise rotation. The position vector of the upper body origin with respect to the world coordinate system $\{W\}$ is $(x_{\text{UB}}, y_{\text{UB}})$.

It is important to distinguish the center of mass (COM) and the geometric center of a segment. The COMs of the body segments, including the thigh, shank, and foot, are determined using experimental data collected by Anderson and Pandy [52]. On the other hand, the COM of the exoskeleton's thigh segment is obtained from SolidWorks (2021, Dassault Systémes

SolidWorks Corp., Waltham, MA, USA) using the mass property tool.

The ground reaction forces (**Fig. 5(C)**) are denoted by $F_{i_1,i_3,i_4}$, where the subscript $i_3 \in \{\text{h},\text{t}\}$ represents heel (h) and toe (t), and $i_4 \in \{\text{n},\text{f}\}$ represents normal force (n) and friction force (f). The detailed forms of the ground reaction forces will be discussed later.

*Equation of motion*

The equation of motion (EOM) of the coupled human-exoskeleton system is derived by means of generalized Lagrange Equation of the second kind:

$$\mathbf{D}(\mathbf{q})\ddot{\mathbf{q}} + \mathbf{C}(\mathbf{q},\dot{\mathbf{q}})\dot{\mathbf{q}} + \mathbf{G}(\mathbf{q}) = \mathbf{Q}, \tag{28}$$

where $\mathbf{q} = (x_{\text{UB}}, y_{\text{UB}}, q_{\text{UB}}, q_{\text{r,h}}, q_{\text{r,k}}, q_{\text{r,a}}, q_{\text{r,e}}, q_{\text{l,h}}, q_{\text{l,k}}, q_{\text{l,a}}, q_{\text{l,e}})^{\text{T}}$ is the generalized coordinate vector, $\mathbf{D}(\mathbf{q}) \in \mathbb{R}^{11 \times 11}$ represents the inertia matrix, $\mathbf{C}(\mathbf{q},\dot{\mathbf{q}}) \in \mathbb{R}^{11 \times 11}$ denotes the matrix for the centrifugal force and Coriolis force, $\mathbf{G}(\mathbf{q}) \in \mathbb{R}^{11 \times 1}$ is the gravity vector, and $\mathbf{Q} \in \mathbb{R}^{11 \times 1}$ represents the generalized force vector, constructed by

$$\mathbf{Q} = \boldsymbol{\tau}_{\text{joint}} + \boldsymbol{\tau}_{\text{exo}} + \boldsymbol{\tau}_{\text{int}} + \mathbf{J}_{\text{grf}}^{\text{T}} \mathbf{F}_{\text{grf}}, \tag{29}$$

where $\boldsymbol{\tau}_{\text{joint}}$, $\boldsymbol{\tau}_{\text{exo}}$, and $\boldsymbol{\tau}_{\text{int}}$ denote the vectors of joint torques, assistive torques generated by the exoskeleton, and the interaction torques between the exoskeleton and human, respectively; $\mathbf{F}_{\text{grf}}$ represents the vector of ground reaction forces, and $\mathbf{J}_{\text{grf}}$ is the Jacobi matrix that transforms the ground reaction forces into the generalized forces. The detailed expressions of the EOM are provided in S2 of the **Supplementary Material**.

*Interaction torques and ground reaction forces*

The interaction torque generated between the exoskeleton and the thigh is characterized by the Kelvin-Voigt model [53], which consists of a linear spring in parallel with a damper. The model describes the relationship between the applied torque and the resulting displacement or velocity:

$$\tau_{i_1,\text{int}} = k_{\text{int}}\, q_{i_1,\text{e}} + d_{\text{int}}\, \dot{q}_{i_1,\text{e}}, \tag{30}$$

where $k_{\text{int}}$ and $d_{\text{int}}$ represents the stiffness and damping coefficients, respectively[54]; $q_{i_1,\text{e}}$ denotes the angle between the exoskeleton's thigh segment and the human thigh.

The normal force, which represents ground reaction force in the direction perpendicular to the ground surface, is characterized by a contact model proposed by Hunt and Crossley [55]. This contact model captures the dynamic interaction between two bodies during a collision and is expressed as

$$F_{i_1,N} = k\delta_{i_1}^{3/2}\left(1 + 1.5 c \dot{\delta}_{i_1}\right), \tag{31}$$

where $k$ represents a stiffness constant incorporating both material and geometry properties,

$c$ is the dissipation coefficient, and $\delta_{i_1}$ represents the depth of contact between the heel/toe and the ground (which remains zero when there is no contact). In our model, both the heel and the toe are modeled as elastic spheres with a radius of 3 cm.

The friction force exerted by the ground on the heel/toe can be expressed as [56]

$$F_{i_1,f} = F_{i_1,N} \left[ \min\left(\frac{v_{i_1}}{v_t}, 1\right) \left( \mu_d + \frac{2(\mu_s - \mu_d)}{1 + (v_{i_1}/v_t)^2} \right) + \mu_v v_{i_1} \right], \quad (32)$$

where $\mu_s$, $\mu_d$, and $\mu_v$ represent the coefficients of static friction, dynamic friction, and viscous friction, respectively; $v_t$ denotes the transition velocity, at which the friction reaches its maximum value; $v_{i_1}$ is the tangential slip velocity between the heel/toe and the ground at the contact point.

So far, we have developed a dynamic model of the coupled human-exoskeleton system that integrates the biological joint torques, exoskeleton assistance, human-exoskeleton interaction, and foot-ground contact. This comprehensive framework allows for the simultaneous consideration of these key factors in the analysis of human-exoskeleton-ground interaction dynamics.

## 3. Model Optimization and Implementation

### 3.1 Model Optimization

The muscle reflex mechanism described in subsection 2.2.1 involves 29 free parameters (i.e., $\mathbf{w} \in \mathbb{R}^{29}$). These parameters enable customization of human motion to adapt to various walking speeds and environmental conditions. Specifically, the optimization problem is constructed as follows

$$\begin{aligned} \min \ & J = J(\mathbf{w}), \ J(\mathbf{w}) = \begin{cases} J_1, & \text{step 1,} \\ J_2, & \text{step 2,} \end{cases} \\ \text{s.t.} \ & F_{\text{opt}}\left(a\mathbf{f}^{\text{L}}(\tilde{l}^{\text{M}})\mathbf{f}^{\text{V}}(\tilde{v}^{\text{M}}) + \beta\tilde{v}^{\text{M}} + \mathbf{f}^{\text{PE}}(\tilde{l}^{\text{M}})\right)\cos\alpha - F_{\text{opt}}\mathbf{f}^{\text{T}}(\tilde{l}^{\text{T}}) = 0, \\ & \mathbf{D}(\mathbf{q})\ddot{\mathbf{q}} + \mathbf{C}(\mathbf{q},\dot{\mathbf{q}})\dot{\mathbf{q}} + \mathbf{G}(\mathbf{q}) = \mathbf{Q}, \\ & 0 \leq \sigma \leq 1, \\ & a_{\min} \leq a \leq 1. \end{aligned} \quad (33)$$

The muscle reflex parameters $\mathbf{w}$ are determined through a two-step optimization process. In step 1, our objective is to maximize the walking distance of the model without falling down; in step 2, we aim to constrain the model's walking behavior utilizing multi-dimensional evaluation criteria so that it satisfies the speed constraint and is more natural and human-like. Throughout both steps, it is essential to satisfy the muscle-tendon force equilibrium constraint (Eq. (18)), the rigid-body dynamic constraint (Eq. (28)), as well as the boundaries of muscle excitation and activation. This approach will be demonstrated to be efficient and robust, eliminating the need for accurate initial guesses.

***Step 1.*** Since the initial guess of the parameters may deviate significantly from their optimum values, the primary objective of this step is to maximize the walking distance achieved by the model within a single simulation. The evaluation of this objective is performed using $J_1$

$$J_1 = \psi - \sqrt{v^2+1} \cdot x_{\text{COM}}(T), \tag{34}$$

where $x_{\text{COM}}$ represents the $x$-coordinate of the model's COM, $T$ denotes the number of timesteps before either the simulation ends or the model falls down, $v$ denotes the tangent value of the ground slope, and $\psi$ is a prescribed parameter indicating the desired distance in meters from the model to the destination.

The optimization switches from step 1 to step 2 when $J_1 < 0$, indicating that the model can walk more than $\psi$ meters in $T$ timesteps. The particle corresponding to this result is then retained and used as the initial guess for optimization in step 2.

***Step 2.*** In this step, the optimization objective is to minimize the following hybrid objective function:

$$J_2 = \omega_{\text{vel}} J_{\text{vel}} + \omega_{\text{angle}} J_{\text{angle}} + \omega_{\text{GRF}} J_{\text{GRF}} + \omega_{\text{effort}} J_{\text{effort}}, \tag{35}$$

where $\omega_{\text{vel}}$, $\omega_{\text{angle}}$, $\omega_{\text{GRF}}$, and $\omega_{\text{effort}}$ are the parameters that weight the four objectives ($J_{\text{vel}}$, $J_{\text{angle}}$, $J_{\text{GRF}}$, and $J_{\text{effort}}$), respectively. The four objective functions are interpreted as follows:

- $J_{\text{vel}}$ represents the velocity fitness, which evaluates the model's ability to achieve the desired velocity at the model's COM ($v_{\text{COM}}^{\text{des}}$), with a significant penalty for falling down:

$$J_{\text{vel}} = \frac{1}{T} \sum_t \left( v_{\text{COM}}(t) - v_{\text{COM}}^{\text{des}} \right)^2 + \delta_{\text{vel}}, \tag{36}$$

where $v_{\text{COM}}(t)$ denotes the velocity of the model's COM at the $t$-th timestep. The first term penalizes the model when its velocity deviates from $v_{\text{COM}}^{\text{des}}$. $\delta_{\text{vel}} = 1$ when the model falls down during the simulation, and $\delta_{\text{vel}} = 0$ otherwise.

- $J_{\text{angle}}$ is the function that constrains ankle hyperflexion and hyperextension

$$J_{\text{angle}} = \frac{1}{T} \sum_t \sum_{i_1=\text{r,l}} \left[ \max\left( q_{i_1,\text{a}}(t) - \hat{q}_{\text{a,max}}, 0 \right) - \min\left( q_{i_1,\text{a}}(t) - \hat{q}_{\text{a,min}}, 0 \right) \right], \tag{37}$$

where $\hat{q}_{\text{a,max}}$ and $\hat{q}_{\text{a,min}}$ are the upper and lower threshold of the ankle motion, respectively. This term equals zero when the ankle motions range within $\left[ \hat{q}_{\text{a,min}}, \hat{q}_{\text{a,max}} \right]$.

- $J_{\text{GRF}}$ imposes a penalty when the normal ground reaction forces exerted on the heels and toes exceed the threshold value $\hat{F}_{\text{GRF}}$:

$$J_{\text{GRF}} = \frac{1}{T}\sum_{t}\frac{1}{mg}\sum_{i_1=r,l}\max\left(F_{i_1,\text{toe,n}}(t) + F_{i_1,\text{heel,n}}(t) - \hat{F}_{\text{GRF}}, 0\right), \tag{38}$$

where $m$ denotes the mass of the model, and $g$ is the gravitational acceleration. This term equals zero when the threshold value is not exceeded.

- $J_{\text{effort}}$ represents the metabolic expenditure of the muscles, following the model developed by Wang et al [57]:

$$J_{\text{effort}} = \frac{1}{x_{\text{COM}}(T)}\sum_{t}\frac{\Delta t_{\text{sim}}}{m}\sum_{i}\left(r_A^i(t) + r_M^i(t) + r_S^i(t) + r_W^i(t)\right), \tag{39}$$

where $x_{\text{COM}}(T)$ denotes the model's walking distance in $T$ timesteps; $\Delta t_{\text{sim}}$ denotes the simulation time step; $r_A$, $r_M$, $r_S$, and $r_W$ denote the rates for muscle activation heat, muscle maintenance heat, muscle shortening heat, and positive mechanical work, respectively (the superscript '$i$' takes TA, SOL, GAS, FEM, HAM, GLU, or ILI ).

To determine the parameters $\mathbf{w}$, the Covariance Matrix Adaptation Evolution Strategy (CMA-ES) [58] is adopted. The CMA-ES optimization is recognized for its effectiveness in solving complex optimization problems, making it a valuable tool in various domains. In the context of CMA-ES optimization, 13 particles are used per iteration, with an initial standard deviation of 0.01. Each particle represents a specific set of free parameters $\mathbf{w}$. In each iteration of the optimization process, the performance of every particle is assessed by the fitness function $J$, based on which, the optimization algorithm adjusts the particles to improve their performances in the subsequent iteration. This iterative process continues until a satisfactory solution is obtained or the maximum number of iterations is reached.

## 3.2 Simulation

We proceed by verifying the feasibility of the dynamic simulation framework and assessing the extent to which the developed model can accurately capture the dynamic responses of the coupled human-exoskeleton system in various scenarios. Our model is designed to be independent of specific walking conditions and environments, enabling a systematic and comprehensive study of their effects. The values of parameters involved in the framework are provided in S3 of the **Supplementary Material**. The proposed framework is implemented and tested in MATLAB (R2021b, The MathWorks Inc., Natick, MA, USA), with a simulation time step of $\Delta t_{\text{sim}} = 5$ ms. The dynamic model, described by ordinary differential equations, is numerically solved using the fourth-order Runge-Kutta method with a step size of 0.5 ms. It is important to note that the primary focus of our simulations lies in verifying the effectiveness and accuracy of the proposed framework in predicting the dynamic responses of the coupled neuromusculoskeletal-exoskeletal system, rather than evaluating the performance of the employed exoskeleton in providing assistance. As a result, our simulations and experiments involve the coupling and interaction between the exoskeleton and the human, without necessarily requiring the exoskeleton to provide assistance.

Simulations are performed under four different scenarios, comprising walking on the flat ground at three walking speed constraints ($v_{COM}^{des} = 0.9$, 1.0, and 1.1 m/s), and walking uphill (5.7° slope (10%)) at $v_{COM}^{des} = 1.0$ m/s. Each scenario is simulated for a duration of 10 seconds using the 2-step optimization process outlined in the previous subsection. In the first step of optimization, we simulate walking on the flat and uphill grounds at a speed of 1.0 m/s to quickly converge to a feasible parameter set. Subsequently, for both ground slope scenarios, we conduct simulations with multiple speed constraints in the second step of optimization, using the outcomes from the first step to refine the parameters. The results reveal that the first step optimization for walking on the flat and uphill grounds requires 33 and 65 iterations, respectively, with an average computing speed of approximately 3 iterations per hour on a workstation (Intel Xeon Platinum 8124M, 3.0 GHz, 128 GB RAM). The second step of optimization requires approximately 19 iterations for each scenario.

**Figure 6** illustrates the simulation results for walking on both flat ground and uphill at a speed of 1 m/s. The developed dynamic simulation framework not only facilitates stable walking that closely resembles human behavior (**Fig. 6(A)**) but also provides a comprehensive understanding of the coupled neuromusculoskeletal-exoskeletal system (**Fig. 6(B-F)**). The obtained results encompass various aspects, including muscle activations and forces (**Fig. 6(B-C)**, exemplified by the HAM, GLU, and SOL muscles of the right leg), joint torques (**Fig. 6(D)**, exemplified by joint torques of the right hip, knee, and ankle joints), human-exoskeleton interaction forces (**Fig. 6(E)**, exemplified by the interaction torque of the right leg), kinematic data (**Fig. 6(F)**, exemplified by the joint angles of the right hip, knee, and ankle), and ground reaction forces (**Fig. 6(G)**). This is particularly significant considering that certain biomechanical information, such as muscle activation, muscle forces, and joint torques, as well as interaction information, such as human-exoskeleton and foot-ground interaction forces, are challenging to measure through instrumentation in experimental settings. An important aspect to highlight is that the obtained results are achieved without relying on experimental data input, leading to substantial savings in experimental costs. As a result, the proposed dynamic model and framework offer a unique capability that opens up the potential for estimating the biomechanics and motion dynamics of humans or human-exoskeleton systems beyond the confines of a laboratory environment, allowing for a broader understanding of the system's behavior and performance in various settings.

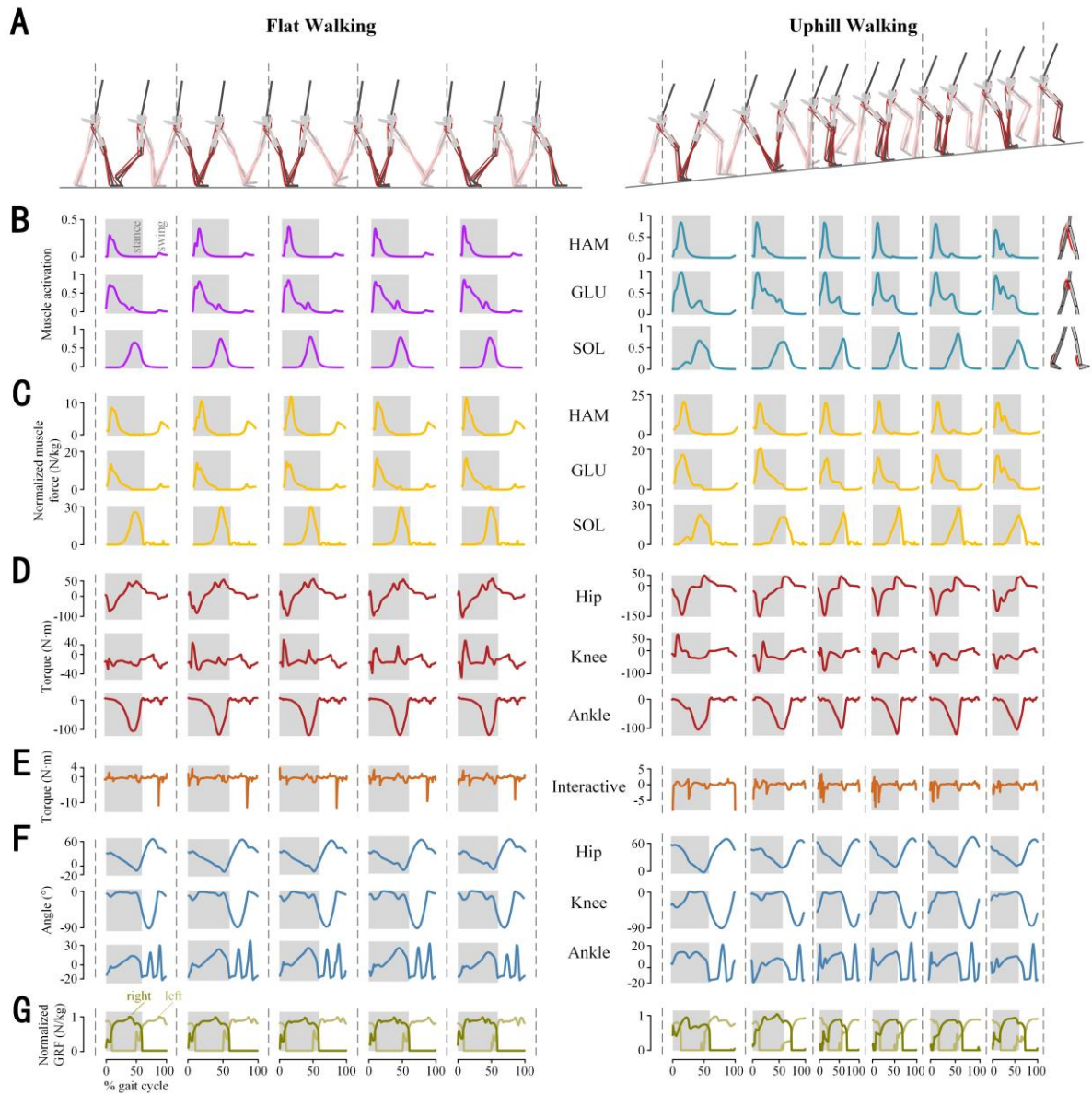

**Fig. 6**. Simulation results of the coupled human-exoskeleton system walking on flat ground and slope. (A) Snapshots of the model walking on flat ground and uphill at a speed of 1 m/s. (B) Normalized muscle activation of HAM, GLU, and SOL of the right leg, where the shaded and unshaded areas denote the stance and swing phases, respectively. (C) Muscle forces of HAM, GLU, and SOL of the right leg, normalized by the subject's body weight. (D) Joint torques of the hip, knee, and ankle of the right leg. (E) Human-exoskeleton interaction torques on the right thigh. (F) Joint angles of the hip, knee, and ankle of the right leg. (G) Ground reaction forces, normalized by the subject's body weight (right and left legs in dark and light green colors, respectively).

The effectiveness of our dynamic simulation framework is further demonstrated by its ability to capture and represent the performance of the human-exoskeleton system under different walking speeds and environments.

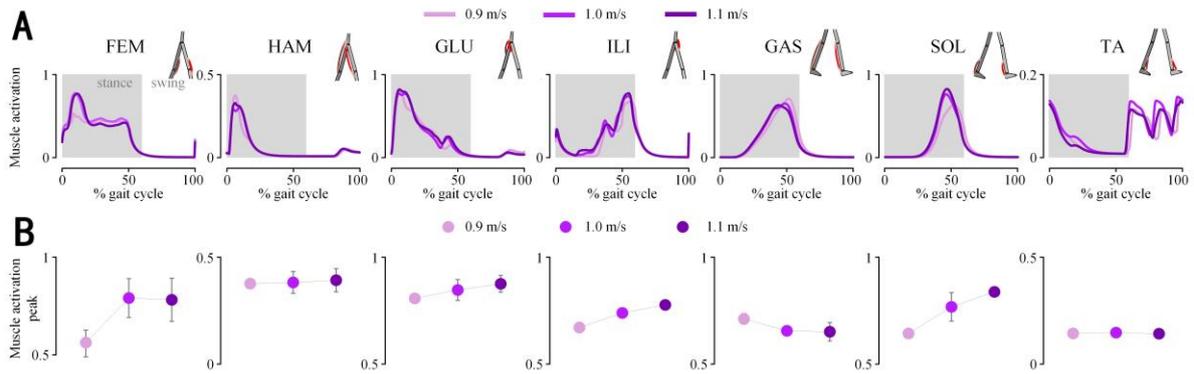

**Fig. 7.** Muscle activations in the right leg with respect to different walking speeds. (A) From the left to the right, showing the muscle activations of the FEM, HAM, GLU, ILI, GAS, SOL, and TA muscles when walking at speeds of 0.9, 1.0, and 1.1 m/s on flat ground. The curves are averaged results of 7 walking cycles. (B) The muscle activation peaks (average ± std) of the respective muscles at the three walking speeds.

Walking speeds are considered as constraints in the optimization problem. Simulations reveal that when the walking speed grows from 0.9 m/s to 1.1 m/s, in addition to increases in step frequency (from 100.44 to 101.81) and in stride length (from 0.5 m to 0.59 m), significant changes in muscle activations occur. **Figure 7** presents the time histories and peak values of muscle activations in the right leg (including the muscles FEM, HAM, GLU, ILI, GAS, SOL, and TA) during a gait cycle for walking on flat ground in relation to different walking speeds (0.9, 1.0, and 1.1 m/s). It is important to note that different muscles exhibit different patterns of change in response to variations in walking speed, which can be attributed to their specific roles within the gait cycle. Specifically, as the walking speed increases from 0.9 to 1.1 m/s, the SOL and ILI muscles, which play crucial roles in propulsion during forward walking, display a positive correlation between peak activation values and walking speeds ($p < 0.05$). The positive correlation between the activation of the FEM muscle and walking speeds can be attributed to its primary role in maintaining knee joint stiffness during the stance phase. This function is vital for supporting the body in an upright position and providing stability during the weight-bearing phase of the gait cycle. As the walking speed increases, there is a greater demand for force generation from the FEM muscle to effectively fulfill this stabilizing role, resulting in the observed positive correlation between its activation and walking speeds. Conversely, it is observed that the peak activation of the GAS muscle decreases with higher walking speeds. This may be attributed to the compensatory behavior from the SOL muscle, which has a similar function to the GAS muscle in the plantar flexor; as the walking speed increases, the activation level of the SOL muscle significantly rises, resulting in a decrease in the activation level of GAS. Notably, the peak activation levels of the GLU and HAM muscles, contributing to gait stability, do not exhibit significant changes with increasing speed. Additionally, the peak activation of the TA muscle remains relatively low and stable as walking speed increases. This phenomenon can be attributed to the influence of stride length on TA muscle activation. This may be due to the tendency in simulations that the foot elevation and clearance is achieved more through the hip and knee joints, thus making the involvement of the FEM and ILI muscles more significant, while the involvement of TA muscle is lower and does not vary significantly relative to walking speeds.

We also demonstrate the capability of our dynamic simulation framework in capturing changes in the environment by analyzing both flat and uphill (5.7° slope) walking conditions at a speed of 1 m/s. Significant changes in stride length and muscle activation peaks are observed on different slopes, highlighting the adaptive nature of our model in adjusting gait according to the environment. Specifically, during uphill walking, the stride length is significantly shortened from 0.58 m of flat walking to 0.44 m, indicating the challenges of walking uphill. Moreover, **Fig. 8** provides a visual representation of the effects of the slope on muscle activations. We observe a significant increase in the activation of GLU and HAM muscles ($p < 0.05$), suggesting the increased effort to maintain gait stability. There is also an increase in GAS muscle activation peaks ($p < 0.05$), reflecting the increased power generation for walking uphill. The lack of significant difference in peak activation of the FEM muscle, as observed in this study and similar to the findings of Saito et al.[59], can be explained by the influence of stride length during uphill walking. The shorter stride length during uphill walking compared to flat walking is advantageous for maintaining stability and support during the stance phase, as it allows for a more controlled and balanced weight transfer. As a result, the demand for force generation from the FEM muscle may not significantly change during uphill walking, leading to a similar peak activation level compared to flat walking conditions. Conversely, the activation peak levels of the SOL, TA, and ILI muscles show significant decreases ($p < 0.05$) during uphill walking. This decline can be primarily attributed to the substantial decreases in stride length, which affects the mechanical demands placed on these muscles. The shorter stride length during uphill walking requires less force generation from the SOL, TA, and ILI muscles for propulsion and support. This decrease in demand is accompanied by compensatory behavior among various muscles. Notably, the SOL and GAS muscles, which share similar functions in plantar flexion, demonstrate compensatory behavior. During uphill walking, the activation level of the GAS muscle notably increases, leading to a corresponding decrease in the activation level of the SOL muscle.

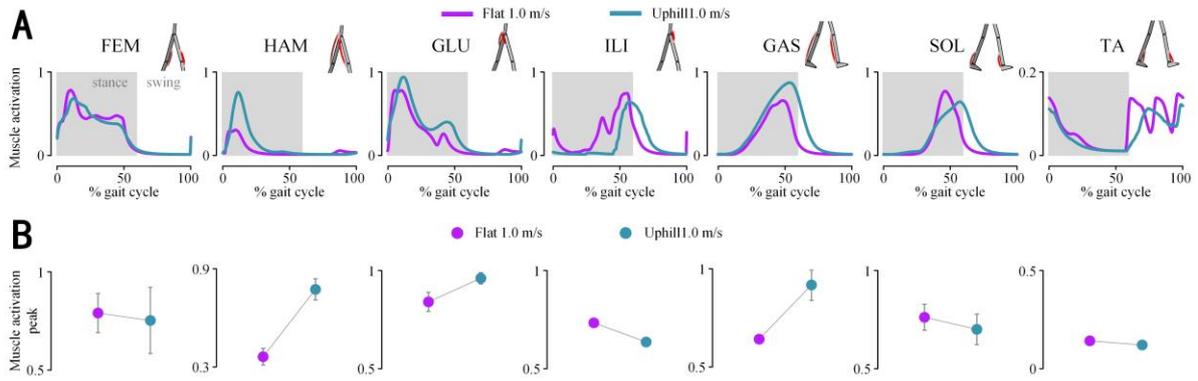

**Fig. 8.** Muscle activations in the right leg with respect to different terrains, the walking speed is 1.0 m/s. (A) From the left to the right, showing the muscle activations of the FEM, HAM, GLU, ILI, GAS, SOL, and TA muscles when walking on flat and uphill (5.7° slope) grounds. The curves are averaged results of 7 walking cycles. (B) The muscle activation peaks (average ± std) of the respective muscles in flat and uphill walking.

It is important to highlight again that our dynamic model and optimization framework do not rely on experimental data, offering us a significant advantage in terms of efficiently integrating and testing updated exoskeleton designs and control strategies in a wide range of scenarios. The results obtained from the framework provide valuable insights into muscle activation patterns and motor control mechanisms during walking across different speeds and environmental conditions. This flexibility in testing and the wealth of information obtained from the framework are instrumental in verifying the adaptability of the system to different environments and different gait patterns. Ultimately, this leads to the provision of optimal assistance tailored to the specific needs of the user.

### 3.3 Experimental Validation

To validate the effectiveness of the dynamic simulation framework and assess the accuracy of the predicted results, we conduct four walking experiments that encompassed walking on the flat ground at three walking speed constraints (0.9, 1.0, and 1.1 m/s), and walking uphill (5.7° slope (10%)) at 1.0 m/s, aligning with the scenarios used in the simulations.

#### 3.3.1 Participant and experiment protocol

A healthy male subject (aged 28, weighing 70 kg, and standing at a height of 171 cm) participates in the experiment. The protocol received approval from the Ethical Committee of Fudan University. The subject is fully informed of the aims and details of the protocol and voluntarily agreed to participate, with the signing of written informed consent.

**Fig. 9(A)** shows the photo of the experimental setup. As the primary objective of the experiment is to validate the effectiveness of the dynamic simulation framework, there is no requirement for the exoskeleton to provide assistance. Hence, the subject involved in the experiment does not carry the exoskeleton's motors and cables, i.e., the subject wears a passive hip exoskeleton that is tightly coupled to the human thigh. To measure the muscle activity, EMG electrodes (Ultium EMG, Noraxon, Scottsdale, AZ, USA) are positioned on five lower limb muscles on both sides: TA, GAS, SOL, FEM, and HAM. The electrodes are firmly secured using medical adhesive tapes to ensure close contact with the skin (**Fig. 9(B)**). Additionally, the subject wears an inertial motion capture suit (Perception Neuron, Noitom, Beijing, China) equipped with multiple inertial sensors. These sensors are attached to elastic straps that are fastened to the corresponding body segments (**Fig. 9(B)**).

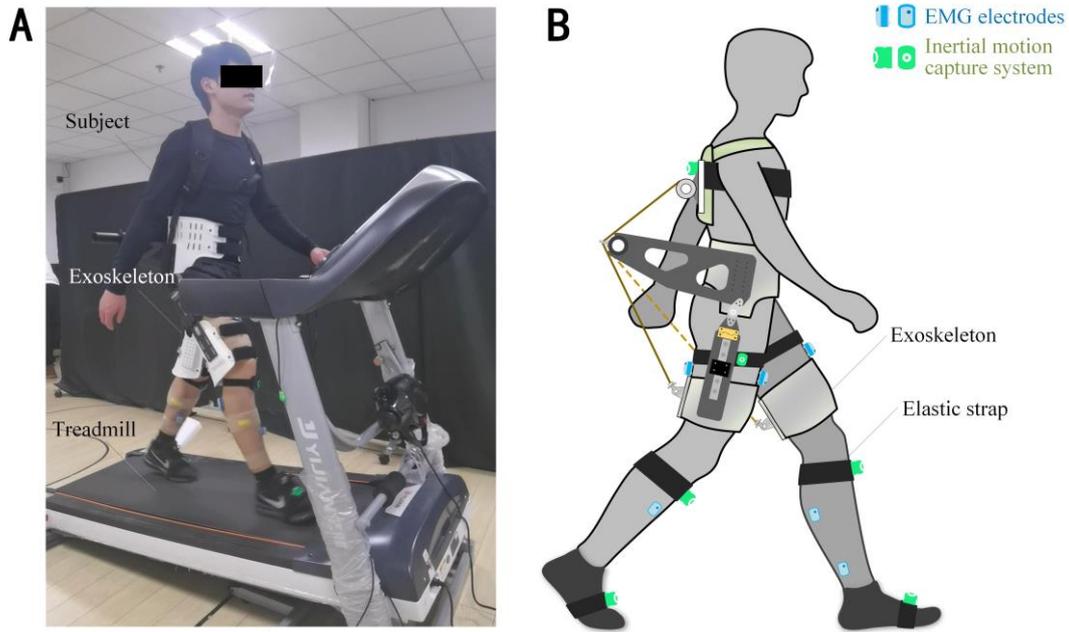

**Fig. 9.** Human walking experiments. (A) The subject wearing the exoskeleton and sensors walks on a treadmill with adjustable slope. (B) Illustration of the sensor placement.

The experimental protocol commences with a pose calibration procedure for the inertial motion capture system. Following the calibration, 2 minutes of still standing is conducted to record the resting surface electromyography (sEMG) signals of the muscles. Subsequently, the subject, wearing the exoskeleton, performs walking tasks on a treadmill under four distinct experimental scenarios. Each scenario has a duration of 1 minute, and a rest period of 30 seconds is provided between scenarios to mitigate muscle fatigue.

### 3.3.2 Data acquisition and process

The sEMG signals of the 10 muscles are measured using the EMG system. The data are processed according to the methodology described in[60]. A sampling frequency of 2000 Hz is utilized for data acquisition. To preprocess the raw sEMG data, a second-order Butterworth high-pass filter with a cut-off frequency of 10 Hz is applied. Subsequently, the data are subjected to full-wave rectification and further filtered using a Butterworth low-pass filter with a cut-off frequency of 5 Hz. To standardize the preprocessed signals, normalization is applied to the sEMG data. The normalization process involves scaling the signals to the peak values observed during walking on flat ground at a speed of 1.0 m/s. This normalization technique ensures that all muscle activation signals are comparable and represents relative changes in muscle activity across different scenarios and walking conditions. The normalization also facilitates meaningful comparison and analysis of sEMG signals and muscle activation during various locomotion tasks in a qualitative manner.

For the measurement of joint angles and the tilt angle of the upper body, an inertial motion capture system is employed. This system provides quaternion measurements of the feet, shanks, thighs, and upper body, enabling the calculation of joint angles of hip, knee, and ankle joints, as well as the tilt angle of the upper body. The sampling frequency of the inertial motion capture system is set to 100 Hz.

### 3.3.3 Experimental results and analysis

To access the capability of the proposed model and optimization framework in replicating human walking behavior, comparisons are conducted between the simulation results and experimental data, as shown in **Fig. 10** and **Fig. 11**. The aim is to validate that the walking motions generated by the simulation framework exhibit similar kinematics and muscle activities to those observed in real experiments.

**Fig. 10(A)** demonstrates the muscle activations generated by the proposed simulation framework at speeds of 0.9, 1.0, and 1.1 m/s, which are compared to the sEMG signals measured during our experiments. It should be noted that sEMG signals and muscle activation are two distinct quantities, and their absolute values cannot be directly compared. Instead, our focus is on observing their patterns of change. Previous gait experiments conducted by Wren et al. [61] have revealed that the coefficients of correlation (R) of sEMG signals among different subjects range from R=0.40 to 0.81. In our study, with the exception of the TA muscles (to be discussed in Section 4), there is a strong consistency (R>0.6375) between muscle activations in simulations and sEMG signals in experiments. These findings provide substantial evidence supporting the effectiveness of the proposed framework in capturing the qualitative neuromuscular reflex patterns exhibited by an exoskeleton-wearing individual during walking.

Furthermore, **Fig. 10(B)** depicts the relationship between peak muscle activity and walking speed in both simulations and experiments. The simulation and experimental data exhibit good agreement in terms of the qualitative variation pattern of peak muscle activity with respect to walking speeds. Specifically, the results indicate that the FEM and SOL muscles demonstrate a notable increase in peak muscle activity as walking speed increases, favorable to support and propel, respectively. The HAM and TA muscles do not exhibit a significant correlation between peak muscle activities and walking speeds. Simulation results indicated a decrease in the peak activation of GAS muscles with increasing walking speed, while the decline in experimental data is not significant. This decrease in peak GAS muscle activity can be attributed to the compensatory behavior between the GAS and SOL muscles.

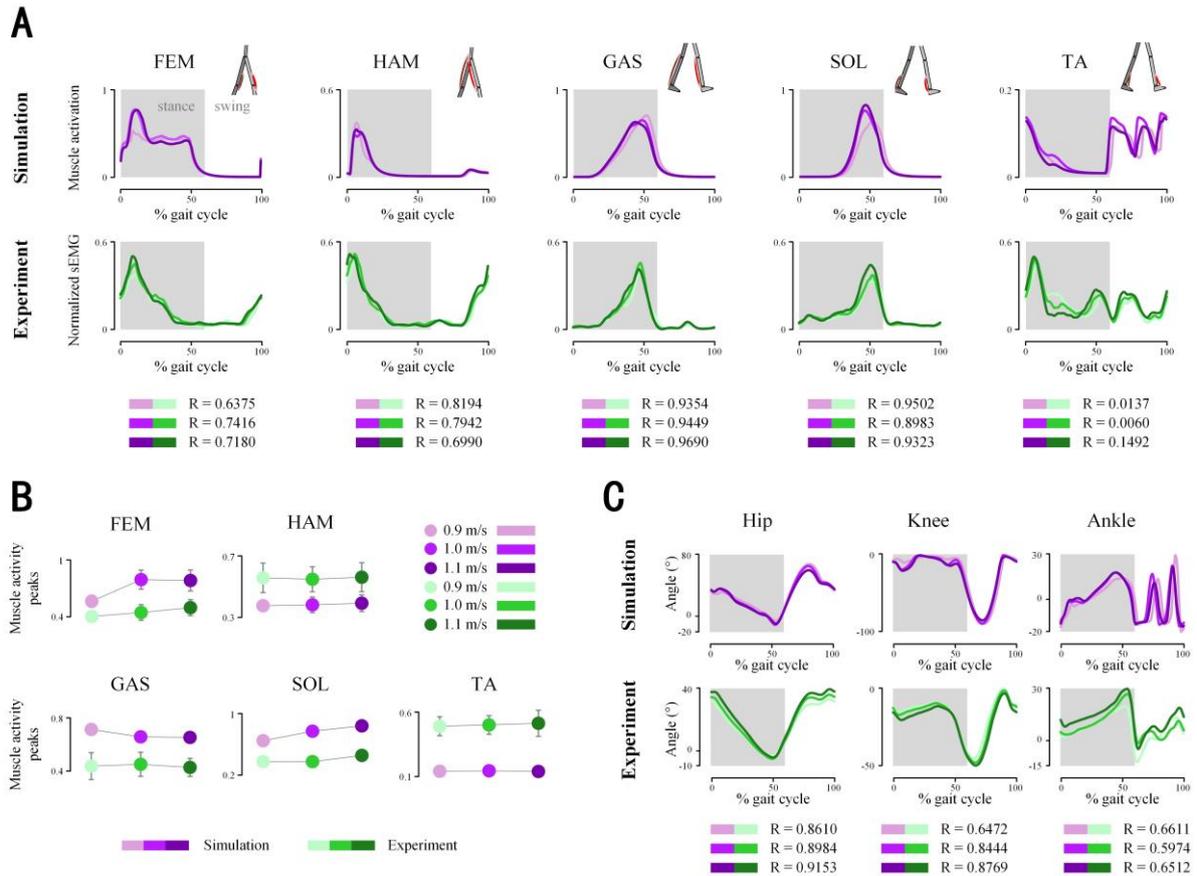

**Fig. 10.** Comparison between simulation and experiment results when walking at speeds of 0.9, 1.0, and 1.1 m/s on flat ground. (A) Activation of FEM, HAM, GAS, SOL, and TA muscles, including both model-predicted muscle activations and experimental sEMG signals. The curves are averaged results of 7 walking cycles, and the coefficients of correlation (R) between the simulated and experimental data are provided. (B) The muscle activation and sEMG peaks (average ± std) of the respective muscles at the three walking speeds. (C) Angles of the hip, knee, and ankle joints, including both model predictions and experimental data captured by inertial motion sensors. The curves are averaged results of 7 walking cycles, and the coefficients of correlation (R) between simulated and experimental data are provided.

In addition, **Fig. 10(C)** demonstrates the consistency between simulation and experiment regarding joint angles (R>0.5974). Note that there are some quantitative discrepancies, particularly in the knee joint angles, which can be attributed to the differences in gait characteristics among individuals. Importantly, our simulation results align with the experimental observations of Wang et al. [33], specifically concerning the knee joint angles.

**Fig. 11** further provides a comprehensive comparison between the simulation and experiment results for walking on flat and uphill terrains. In **Fig. 11(A)**, it is evident that, with the exception of TA muscles (to be discussed in Section 4), the predicted muscle activation in the simulation aligns well with the measured sEMG signals in terms of the qualitative trends, with R>0.4749. **Fig. 11(B)** illustrates the effects of slope on peak muscle activities in both simulations and experiments. The patterns of change in peak muscle activity relative to the terrain are qualitatively consistent in both simulation and experimental results. Specifically, the FEM muscle does not exhibit a significant correlation between the peak muscle activity and slope, because it is mainly responsible for support. The GAS and HAM muscles receive significant increases in peak muscle activities when walking uphill to increase propulsion and enhance stability, respectively. On the contrary, there is a slight decrease in peak muscle activity of the TA muscles during uphill walking. Note that the peak values of sEMG of the SOL muscle increase significantly in experiments during uphill walking, while there is a slight decrease in simulations. This discrepancy may be attributed to the fact that the GAS and SOL muscles have different compensatory behavior in simulations and experiments.

Moreover, **Fig. 11(C)** reveals that, except for the knee joint angle during uphill walking, the simulated and experimental joint angles are qualitatively consistent in terms of their variation trends (R>0.4279). The lack of correlation in knee joint angles during uphill walking may be due to the delay in the maximum knee flexion angle between simulations and experiments. The observed delay in knee flexion during uphill walking can be attributed to the optimization of the gait pattern based on energy optimization criteria. In order to adapt to the uphill terrain, the simulation framework optimizes the gait pattern by adjusting various muscle activations and joint angles. This optimization process prioritizes energy efficiency, which may lead to a delay in achieving the maximum knee flexion angle compared to the experimental results. Although incorporating reference kinematics into optimization objectives can enhance the fitting quality for a specific task, the introduction of experimental data can significantly limit the flexibility of the simulation framework. The reliance on experimental data would restrict the simulation framework's ability to explore alternative solutions or adapt to different scenarios.

Overall, the comparison between experimental and simulation results fully demonstrates the ability of our dynamic model and simulation framework to accurately predict muscle activity and kinematics. Importantly, our model is able to generate rich and informative outputs without relying on experimental tests or sensor data. This feature instills confidence in utilizing this dynamics simulation framework for studying the neuromusculoskeletal dynamics of the coupled human-exoskeleton system across various environments and gait patterns.

**Fig. 11.** Comparison between simulation and experiment results when walking flat and uphill at a speed of 1.0 m/s. (A) Activation of FEM, HAM, GAS, SOL, and TA muscles, including both model-predicted muscle activations and experimental sEMG signals. The curves are averaged results of 7 walking cycles, and the coefficients of correlation (R) between simulated and experimental data are provided. (B) The muscle activation and sEMG peaks (average ± std) of the respective muscles when walking on the two terrains. (C) Angles of the hip, knee, and ankle joints, including both model predictions and experimental data captured by inertial motion sensors. The curves are averaged results of 7 walking cycles, and the coefficients of correlation (R) between simulated and experimental data are provided.

## 4. Discussion

This study presents an optimization-based comprehensive dynamic simulation framework for the thorough analysis of human-exoskeleton coupling systems. The framework integrates the human neuromusculoskeletal system with exoskeletons and incorporates adjustable human-exoskeleton coupling and foot-ground contact. It enables the prediction and analysis of joint kinematics of the coupled exoskeleton-human system in various walking conditions and environments, while also integrating a complete neuromusculoskeletal feedback loop for extensive biomechanical analysis. The introduced simulation framework possesses the unique capability to comprehensively simulate the multibody dynamics of the coupled human-exoskeleton system in various scenarios without relying on experimental data. This eliminates the need for frequent production of new exoskeleton prototypes and expensive human experiments, while enabling efficient and cost-effective performance testing of novel or improved exoskeleton designs and control strategies. The proposed simulation framework also offers valuable insights into the interaction forces between humans and exoskeletons, which are crucial for assessing the effectiveness and safety of exoskeletons in enhancing human movement. However, it is important to acknowledge certain limitations and minor discrepancies in the quantitative aspect of some results. The succeeding subsections will offer a detailed discussion of these topics by delving into their specifics.

**4.1 Advantages and limitations**

Notably, the proposed simulation framework employs an optimization-based approach to determine multi-source biomechanical signals, including muscle activation, muscle forces, joint torques, and joint kinematics, without relying on experimental data. Rather than utilizing empirical preset values or direct optimization, these signals are obtained by optimizing the muscle reflex parameters, considering various interactions such as muscle activities, muscle contraction dynamics, human-exoskeleton interaction, and system-environment interaction. This approach, along with improved segmentation of the gait phase, significantly improves the accuracy of muscle excitation, enhances the representation of physiological behavior, and offers a realistic depiction of human movement within the dynamic simulation context.

The simulation of the coupled neuromusculoskeletal-exoskeletal system in this framework is achieved through a well-designed optimization strategy. The optimization process consists of two steps, each serving a specific purpose. In the first step, our primary goal is to maximize the walking distance without falls. The focus is on ensuring that the model exhibits reasonable and feasible walking behavior. In the second step, a multidimensional evaluation criterion is employed, incorporating factors such as walking speed, joint angle constraints, and muscle energy consumption. This criterion is employed to guide the optimization process and constrain the model's walking behavior, aiming for a more natural and human-like gait pattern. This step-wise optimization approach offers effectiveness and robustness, facilitating the convergence of the optimization even when accurate initial guesses are not available.

The developed simulation framework in this study excels in accurately capturing the dynamic response of the coupled human-exoskeleton system across a range of walking speeds and environmental conditions. This can be attributed to the incorporation of a neuro-musculoskeletal closed loop, which is achieved through the optimization of muscle reflex parameters using multidimensional evaluation criteria. The optimization process ensures that important constraints related to muscle-tendon force balance, rigid body dynamics, joint angle limits, and the boundaries of muscle excitation and activation are consistently met. Moreover, the variations in environment and walking speed are appropriately accounted for in the optimization objectives and the adjustable foot-ground contact forces within the rigid body dynamics. These factors play a key role in shaping the neuromusculoskeletal closed-loop and subsequently influence the kinematic and biomechanical outcomes of the coupled system.

While our method offers several notable advantages, it is important to acknowledge the limitations of this study. One limitation is the relatively low computational efficiency of the simulation framework, albeit a two-step optimization strategy and parallel optimization with the CMA-ES algorithm have been used. This is an important consideration, as it may limit the applicability of the framework in certain real-time or time-sensitive applications. To address this limitation and improve computational efficiency, further research and development efforts could explore alternative optimization algorithms or parallel computing techniques, which could help reduce the computational burden and enable faster simulations without compromising the accuracy and reliability of the results. Efforts to enhance computational efficiency would not only make the simulation framework more practical for time-sensitive applications but also expand its potential applications in areas such as real-time control and virtual reality simulations.

Another limitation of this study is the presence of individual differences in gait characteristics and the difficulty in accurately measuring certain parameters of the human muscle model. These factors can induce quantitative differences between the simulation results and experimental data. While exact numerical agreement between the simulation and experimental results cannot always be met or pursued, the overall behavior and trends exhibited by the system remain consistent. This qualitative consistency allows for valuable insights into the dynamics and interactions of the human-exoskeleton system, aiding in understanding its performance and improving its design. In addition, the sample size used in this study is limited to a single participant who performed gait analysis repeatedly. Future studies may benefit from a larger sample size and more extensive experimental validation involving diverse populations performing various activities in different environments. This would help validate and refine the modeling and simulation approaches, further demonstrating the accuracy and applicability of the framework.

Lastly, it should be noted that our optimization method does not provide a guarantee of convergence to the global optimum. Instead, it may converge to different local optima depending on the specific problem and initial conditions. However, after a sufficient number of iterations, our optimization method can eventually converge to a similar local optimal solution that effectively characterizes human motion within the constraints and objectives defined in the optimization problem. While it may not be the absolute global optimum, the obtained solution captures the essential characteristics and behaviors of human movement. To mitigate the impact of local optima, it is crucial to carefully design the optimization problem by incorporating appropriate constraints, objectives, and evaluation criteria. Additionally, conducting sensitivity analyses and exploring different optimization algorithms may further enhance the convergence behavior and robustness of the optimization process.

**4.2 Result interpretation and error analysis**

Using the dynamic simulation framework proposed in this study, we have conducted a comprehensive set of simulations and experiments to analyze the gait of the coupled human-exoskeleton system under different terrains and walking speeds. The results obtained from these investigations demonstrate that the proposed framework successfully captures the patterns of joint kinematics and muscle activity during walking across a wide range of speeds and environmental conditions. Notably, the simulated patterns closely align with the trends observed in experimental data, providing strong evidence for the effectiveness and accuracy of the dynamic simulation framework. This validation highlights the framework's capability to faithfully reproduce and predict the behavior of the human-exoskeleton system during walking, establishing it as a valuable tool for further research in this field.

The simulation framework presented in this study can accurately capture the functional variations in muscle activity during different walking conditions. Specifically, it effectively represents the dynamic changes in the activity of different muscles based on their specific functions. For instance, the GLU and HAM muscles, which contribute to walking stability, exhibit no significant correlation between peak muscle activities and walking speeds. However, during uphill walking, there is a notable increase in the peak activity of these muscles, indicating their role in providing stability in changing terrains. The FEM muscle, responsible for support, shows a positive correlation with walking speed, reflecting its involvement in generating the necessary support for faster walking. Interestingly, during uphill walking, where a shorter stride length promotes stability, no significant correlation is observed. Moreover, the proposed framework successfully captures the phenomenon of muscle compensation between muscles with similar functions, as evidenced by the interactions between the SOL and GAS muscles. This ability to represent such muscle interactions and compensatory mechanisms is achieved by incorporating a closed neuromusculoskeletal feedback loop, coupled with a rigid-body dynamics model of the human-exoskeleton system. The solution of this coupled system is obtained by optimization, considering factors such as velocity, stability, energy, and other objectives. While the obtained solution may not be globally optimal, it physiologically demonstrates the variation pattern of different muscle activities relative to walking conditions.

While the proposed framework demonstrates its effectiveness in qualitatively capturing the kinematics and biomechanics during walking with exoskeletons, it is important to acknowledge the existence of minor quantitative disparities between the simulation results and the experimental data, particularly with regard to the activity of the TA muscle. These disparities can be attributed to several factors, including individual differences in gait characteristics and limitations in the modeling of the muscle's driving mechanism in the simulation. In our model, the excitation of the TA muscle is determined by the positive feedback of the muscle length and the negative feedback of the activity of the SOL muscle. However, the actual function of the TA muscle during walking is more complex and multifaceted. In reality, the TA muscle serves as the primary dorsiflexor of the ankle joint, requiring a larger force generation after heel strike to maintain the ankle joint in a neutral position and prevent foot slapping. Additionally, it needs to contract during the swing phase to ensure proper foot clearance. These dynamic functions of the TA muscle are not fully captured in our current model, leading to noticeable differences between the simulated and actual TA muscle activities. Furthermore, the synergistic and antagonistic interactions between TA muscle and other muscles involved in ankle joint control may be more intricate than our current understanding. The complexity of these interactions and the precise coordination among muscles during the walking process present challenges in accurately modeling and simulating their behavior.

While the proposed framework provides valuable insights into the overall dynamics of the human-exoskeleton system and captures qualitative trends in muscle activity, the discrepancies observed in the TA muscle activity highlight the need for further refinement and development of the muscle modeling approach. Future research efforts can focus on incorporating more detailed and comprehensive muscle models that better capture the intricate dynamics and interactions among muscles, leading to a more accurate representation of muscle activity during walking with exoskeletons.

## 5. Conclusions

This paper presents a comprehensive multi-body dynamic simulation framework that enables the prediction and analysis of the interactions and performance of coupled neuromusculoskeletal-exoskeletal systems across various environments. The simulation framework encompasses essential components including human lower limb muscle reflex mechanisms, muscle activation dynamics, muscle-tendon contraction dynamics, musculoskeletal dynamics, human body-exoskeleton interaction, and foot-ground contact. Notably, the simulation framework does not rely on external measurements or empirical data but instead employs an optimization-based approach that considers multi-objective fitness functions. This allows for a thorough analysis of the kinematic and biomechanical characteristics of the coupled human-exoskeleton system under diverse walking and environmental conditions. By comparing the simulation results with experimental data, the study demonstrates the convenience, effectiveness, and robustness of the proposed framework in capturing the qualitative neuromuscular reflex patterns and joint kinematics observed in individuals wearing exoskeletons during walking across different scenarios. While certain limitations exist, they pave the way for future research directions in the dynamic simulation of human-exoskeleton coupling.

Furthermore, the model-based simulation framework presented in this study provides a wealth of information that facilitates efficient and cost-effective performance testing of the human-exoskeleton system under various environmental and gait conditions. This capability is particularly advantageous for integrating and evaluating novel or updated exoskeleton designs and control strategies across a diverse range of scenarios. By eliminating the need for frequent prototyping of new exoskeletons and conducting expensive human experiments, the simulation framework would streamline the development and evaluation process, leading to significant time and cost savings.

## Acknowledgments

This research was supported by the National Natural Science Foundation of China (Grants No. 12272096) and the Shanghai Pilot Program for Basic Research - Fudan University 21TQ1400100-22TQ009.

Supplementary Material for

# A Comprehensive Dynamic Simulation Framework for Coupled Neuromusculoskeletal-Exoskeletal Systems

Wei Jin[1,2,#], Jiaqi Liu[3,#], Qiwei Zhang[1,2], Xiaoxu Zhang[1,2], Qining Wang[4], Hongbin Fang[1,2,*], Jian Xu[3]

[1] Institute of AI and Robotics, State Key Laboratory of Medical Neurobiology, MOE Engineering Research Center of AI & Robotics, Fudan University, Shanghai 200433, China

[2] Yiwu Research Institute, Fudan University, Yiwu, Zhejiang Province 322000, China

[3] School of Aerospace Engineering and Applied Mechanics, Tongji University, Shanghai 200092, China

[4] College of Engineering, Peking University, Beijing 100871, China

[#] W.J. and J.L. contributed equally to this work.

[*] To whom correspondence should be addressed. Email: fanghongbin@fudan.edu.cn (H.F.)

# S1 Musculotendon contraction velocity derivation

The muscle-tendon force equilibrium is

$$F_{opt}\left(a\mathbf{f}^{L}(\tilde{l}^{M})\mathbf{f}^{V}(\tilde{v}^{M})\beta\tilde{v}^{M}+\mathbf{f}^{PE}(\tilde{l}^{M})\right)\cos\alpha - F_{opt}\mathbf{f}^{T}(\tilde{l}^{T}) = 0, \tag{S1}$$

where $F_{opt}$ is the optimal isometric force, $a$ is the muscle activation, and $\beta$ is the damping coefficient. $\alpha$ is the pennation angle of the muscle. $\tilde{l}^{M}$ is the muscle length normalized by its optimal values, i.e., $\tilde{l}^{M} = l^{M}/l^{M}_{opt}$; $\tilde{l}^{T}$ is the tendon length normalized by its slack length, i.e., $\tilde{l}^{T} = l^{T}/l^{T}_{S}$. $\tilde{v}^{M}$ is the derivative of $\tilde{l}^{M}$ with respect to time, i.e., $\tilde{v}^{M} = \dot{\tilde{l}}^{M}$. $\mathbf{f}^{L}(\tilde{l}^{M})$ depicts the relationship between the active force and muscle length, $\mathbf{f}^{V}(\tilde{v}^{M})$ depicts the relationship between the active force and the muscle contraction velocity, and $\mathbf{f}^{T}(\tilde{l}^{T})$ represents the relationship between the tendon force and tendon length.

Hence, The muscle contraction velocity yields

$$\tilde{v}^{M} = \mathbf{f}^{V}_{inv}\left(\frac{\mathbf{f}^{T}(\tilde{l}^{T})/\cos\alpha - \beta\tilde{v}^{M} - \mathbf{f}^{PE}(\tilde{l}^{M})}{a\mathbf{f}^{L}(\tilde{l}^{M})}\right), \tag{S2}$$

where $\mathbf{f}^{V}_{inv}$ is the inverse function of $\mathbf{f}^{V}$. Unfortunately, $\tilde{v}^{M}$ cannot be calculated directly through Eq. (S2) as it appears on both sides of the equation. As a substitution, we utilized Newton's method to approach the true value $\tilde{v}^{M}$. According to Eq. (S1), the residual force $\varepsilon$, i.e. the difference between the tendon force and the muscle force along the tendon is

$$\varepsilon = F_{opt}\left(a\mathbf{f}^{L}(\tilde{l}^{M})\mathbf{f}^{V}(\tilde{v}^{M}) + \beta\tilde{v}^{M} + \mathbf{f}^{PE}(\tilde{l}^{M})\right)\cos\alpha - F_{opt}\mathbf{f}^{T}(\tilde{l}^{T}). \tag{S3}$$

Then the derivative of $\varepsilon$ with respect to $\tilde{v}^{M}$ is

$$\begin{aligned}\varepsilon' &= \frac{d\varepsilon}{d\tilde{v}^{M}} \\ &= \frac{d}{d\tilde{v}^{M}}\left(F_{opt}\left(a\,\mathbf{f}^{L}(\tilde{l}^{M})\mathbf{f}^{V}(\tilde{v}^{M}) + \beta\tilde{v}^{M} + \mathbf{f}^{PE}(\tilde{l}^{M})\right)\cos\alpha - F_{opt}\mathbf{f}^{T}(\tilde{l}^{T})\right) \\ &= a\,F_{opt}\frac{d}{d\tilde{v}^{M}}\left(\mathbf{f}^{L}(\tilde{l}^{M})\mathbf{f}^{V}(\tilde{v}^{M}) + \beta\tilde{v}^{M}\right)\cos\alpha \\ &= a\,F_{opt}\left(\mathbf{f}^{L}(\tilde{l}^{M})\frac{d(\mathbf{f}^{V}(\tilde{v}^{M}))}{d\tilde{v}^{M}} + \beta\right)\cos\alpha.\end{aligned} \tag{S4}$$

So the new approximation $\tilde{v}^{M}$ is

$$\tilde{v}^{M} = \tilde{v}^{M} - \frac{\varepsilon}{\varepsilon'}. \tag{S5}$$

Eq. (S3)–(S5) can be iteratively operated upon until the residual force reaches the threshold:

$$\varepsilon < \varepsilon_{0}. \tag{S6}$$

## S2 Multi-rigid-body dynamics

The Lagrange equation of the second kind can be written as

$$\frac{\mathrm{d}}{\mathrm{d}t}\left(\frac{\partial L}{\partial \dot{\mathbf{q}}}\right) - \frac{\partial L}{\partial \mathbf{q}} = \mathbf{Q}, \tag{S7}$$

$$L = T - V, \tag{S8}$$

where $L$ represents the Lagrangian, $\mathbf{Q}$ is the generalized force vector, and $\mathbf{q}$ is the generalized coordinate vector, selected as

$$\mathbf{q} = \left(x_{\mathrm{UB}}, y_{\mathrm{UB}}, q_{\mathrm{UB}}, q_{\mathrm{r,h}}, q_{\mathrm{r,k}}, q_{\mathrm{r,a}}, q_{\mathrm{r,e}}, q_{\mathrm{l,h}}, q_{\mathrm{l,k}}, q_{\mathrm{l,a}}, q_{\mathrm{l,e}}\right)^{\mathrm{T}}.$$

$T$ and $V$ are the kinetic energy and potential energy of the system, respectively.

The kinetic energy of the $j$-th segment is

$$T_j = \frac{1}{2} I_j \dot{\theta}_j^2 + \frac{1}{2} m_j \left(\dot{x}_j^2 + \dot{y}_j^2\right), \tag{S9}$$

where $I_j$ and $m_j$ are the moment of inertia and mass of the $j$-th segment, respectively. The position vector of COM of the segment with respect to $\{\mathrm{W}\}$ is $(x_j, y_j)$; $\theta_j$ denotes the absolute angle of the $j$-th segment. $j = 1$–$9$ represents the upper body, right thigh, right shank, right foot, right thigh segment of the exoskeleton, left thigh, left shank, left foot, and left thigh segment of the exoskeleton, respectively.

Then the total kinetic energy of the system is

$$T = \frac{1}{2}\sum_j T_j = \frac{1}{2}\sum_j \left[I_j \dot{\theta}_j^2 + m_j \left(\dot{x}_j^2 + \dot{y}_j^2\right)\right]. \tag{S10}$$

The potential energy of the $j$-th segment is

$$V_j = m_j g\, y_j, \tag{S11}$$

where $g$ denotes the gravitational acceleration. The total potential energy of the system is

$$V = \sum_j V_j = \sum_j m_j g\, y_j. \tag{S12}$$

Substitute Eq. (S8) into Eq. (S7), we have

$$\frac{\mathrm{d}}{\mathrm{d}t}\left(\frac{\partial T}{\partial \dot{\mathbf{q}}}\right) - \frac{\mathrm{d}}{\mathrm{d}t}\left(\frac{\partial V}{\partial \dot{\mathbf{q}}}\right) - \frac{\partial T}{\partial \mathbf{q}} + \frac{\partial V}{\partial \mathbf{q}} = \mathbf{Q}. \tag{S13}$$

For the first term on the left-hand side:

$$\frac{\partial T}{\partial \dot{\mathbf{q}}} = \frac{1}{2} \sum_j \left[ I_j \frac{\partial (\dot{\theta}_j^2)}{\partial \dot{\mathbf{q}}} + m_j \frac{\partial (\dot{x}_j^2 + \dot{y}_j^2)}{\partial \dot{\mathbf{q}}} \right]$$

$$= \sum_j \left[ I_j \frac{\partial \dot{\theta}_j}{\partial \dot{\mathbf{q}}} \dot{\theta}_j + m_j \frac{\partial \dot{x}_j}{\partial \dot{\mathbf{q}}} \dot{x}_j + m_j \frac{\partial \dot{y}_j}{\partial \dot{\mathbf{q}}} \dot{y}_j \right]$$

$$= \sum_j \left[ \frac{\partial \theta_j}{\partial \mathbf{q}} I_j \left(\frac{\partial \theta_j}{\partial \mathbf{q}}\right)^{\mathrm{T}} \dot{\mathbf{q}} + \frac{\partial x_j}{\partial \mathbf{q}} m_j \left(\frac{\partial x_j}{\partial \mathbf{q}}\right)^{\mathrm{T}} \dot{\mathbf{q}} + \frac{\partial y_j}{\partial \mathbf{q}} m_j \left(\frac{\partial y_j}{\partial \mathbf{q}}\right)^{\mathrm{T}} \dot{\mathbf{q}} \right] \quad \text{(S14)}$$

$$= \sum_j \left[ \frac{\partial \theta_j}{\partial \mathbf{q}} I_j \left(\frac{\partial \theta_j}{\partial \mathbf{q}}\right)^{\mathrm{T}} + \frac{\partial x_j}{\partial \mathbf{q}} m_j \left(\frac{\partial x_j}{\partial \mathbf{q}}\right)^{\mathrm{T}} + \frac{\partial y_j}{\partial \mathbf{q}} m_j \left(\frac{\partial y_j}{\partial \mathbf{q}}\right)^{\mathrm{T}} \right] \dot{\mathbf{q}},$$

then

$$\frac{\mathrm{d}}{\mathrm{d}t} \left( \frac{\partial T}{\partial \dot{\mathbf{q}}} \right) = \sum_j \left[ \frac{\partial \theta_j}{\partial \mathbf{q}} I_j \left(\frac{\partial \theta_j}{\partial \mathbf{q}}\right)^{\mathrm{T}} + \frac{\partial x_j}{\partial \mathbf{q}} m_j \left(\frac{\partial x_j}{\partial \mathbf{q}}\right)^{\mathrm{T}} + \frac{\partial y_j}{\partial \mathbf{q}} m_j \left(\frac{\partial y_j}{\partial \mathbf{q}}\right)^{\mathrm{T}} \right] \ddot{\mathbf{q}}$$

$$+ \sum_j \left[ \frac{\mathrm{d}}{\mathrm{d}t}\left(\frac{\partial x_j}{\partial \mathbf{q}}\right) m_j \left(\frac{\partial x_j}{\partial \mathbf{q}}\right)^{\mathrm{T}} + \frac{\partial x_j}{\partial \mathbf{q}} m_j \frac{\mathrm{d}}{\mathrm{d}t}\left(\frac{\partial x_j}{\partial \mathbf{q}}\right)^{\mathrm{T}} + \frac{\mathrm{d}}{\mathrm{d}t}\left(\frac{\partial y_j}{\partial \mathbf{q}}\right) m_j \left(\frac{\partial y_j}{\partial \mathbf{q}}\right)^{\mathrm{T}} \right.$$

$$\left. + \frac{\partial y_j}{\partial \mathbf{q}} m_j \frac{\mathrm{d}}{\mathrm{d}t}\left(\frac{\partial y_j}{\partial \mathbf{q}}\right)^{\mathrm{T}} \right] \dot{\mathbf{q}}$$

$$= \sum_j \left[ \frac{\partial \theta_j}{\partial \mathbf{q}} I_j \left(\frac{\partial \theta_j}{\partial \mathbf{q}}\right)^{\mathrm{T}} + \frac{\partial x_j}{\partial \mathbf{q}} m_j \left(\frac{\partial x_j}{\partial \mathbf{q}}\right)^{\mathrm{T}} + \frac{\partial y_j}{\partial \mathbf{q}} m_j \left(\frac{\partial y_j}{\partial \mathbf{q}}\right)^{\mathrm{T}} \right] \ddot{\mathbf{q}}$$

$$+ \sum_j \left[ \frac{\partial \dot{x}_j}{\partial \mathbf{q}} m_j \left(\frac{\partial x_j}{\partial \mathbf{q}}\right)^{\mathrm{T}} + \frac{\partial x_j}{\partial \mathbf{q}} m_j \left(\frac{\partial \dot{x}_j}{\partial \mathbf{q}}\right)^{\mathrm{T}} + \frac{\partial \dot{y}_j}{\partial \mathbf{q}} m_j \left(\frac{\partial y_j}{\partial \mathbf{q}}\right)^{\mathrm{T}} \right.$$

$$\left. + \frac{\partial y_j}{\partial \mathbf{q}} m_j \left(\frac{\partial \dot{y}_j}{\partial \mathbf{q}}\right)^{\mathrm{T}} \right] \dot{\mathbf{q}}$$

$$= \left( \mathbf{I} + \mathbf{J}_{\mathbf{x}_{\mathrm{seg}}}^{\mathrm{T}} \mathbf{M} \mathbf{J}_{\mathbf{x}_{\mathrm{seg}}} \right) \ddot{\mathbf{q}} + \left( \dot{\mathbf{J}}_{\mathbf{x}_{\mathrm{seg}}}^{\mathrm{T}} \mathbf{M} \mathbf{J}_{\mathbf{x}_{\mathrm{seg}}} + \mathbf{J}_{\mathbf{x}_{\mathrm{seg}}}^{\mathrm{T}} \mathbf{M} \dot{\mathbf{J}}_{\mathbf{x}_{\mathrm{seg}}} \right) \dot{\mathbf{q}}, \quad \text{(S15)}$$

where $\mathbf{I}$ is the inertial matrix:

$$\mathbf{I} = \begin{bmatrix} 0 & 0 & 0 & 0 & 0 & 0 & 0 & 0 & 0 & 0 & 0 \\ 0 & 0 & 0 & 0 & 0 & 0 & 0 & 0 & 0 & 0 & 0 \\ 0 & 0 & \sum_j I_j & I_2+I_3+I_4+I_5 & I_3+I_4 & I_4 & I_5 & I_6+I_7+I_8+I_9 & I_7+I_8 & I_8 & I_9 \\ 0 & 0 & I_2+I_3+I_4+I_5 & I_2+I_3+I_4+I_5 & I_3+I_4 & I_4 & I_5 & 0 & 0 & 0 & 0 \\ 0 & 0 & I_3+I_4 & I_3+I_4 & I_3+I_4 & I_4 & 0 & 0 & 0 & 0 & 0 \\ 0 & 0 & I_4 & I_4 & I_4 & I_4 & 0 & 0 & 0 & 0 & 0 \\ 0 & 0 & I_5 & I_5 & 0 & 0 & I_5 & 0 & 0 & 0 & 0 \\ 0 & 0 & I_6+I_7+I_8+I_9 & 0 & 0 & 0 & 0 & I_6+I_7+I_8+I_9 & I_7+I_8 & I_8 & I_9 \\ 0 & 0 & I_7+I_8 & 0 & 0 & 0 & 0 & I_7+I_8 & I_7+I_8 & I_8 & 0 \\ 0 & 0 & I_8 & 0 & 0 & 0 & 0 & I_8 & I_8 & I_8 & 0 \\ 0 & 0 & I_9 & 0 & 0 & 0 & 0 & I_9 & 0 & 0 & I_9 \end{bmatrix}, \quad (S16)$$

$\mathbf{M}$ is the diagonal mass matrix:

$$\mathbf{M} = \mathrm{diag}\{m_1, m_2, \ldots, m_9, m_1, m_2, \ldots, m_9\}, \quad (S17)$$

and $\mathbf{J}_{\mathbf{x}_{seg}}$ is the Jacobi matrix:

$$\mathbf{J}_{\mathbf{x}_{seg}} = \frac{\partial \mathbf{x}_{seg}}{\partial \mathbf{q}^{\mathrm{T}}}, \quad (S18)$$

$$\mathbf{x}_{seg} = [x_1, x_2, \ldots, x_9, y_1, y_2, \ldots, y_9]^{\mathrm{T}}. \quad (S19)$$

For the second term on the left hand side:

$$\frac{\mathrm{d}}{\mathrm{d}t}\left(\frac{\partial V}{\partial \dot{\mathbf{q}}}\right) = 0. \quad (S20)$$

For the third term on the left-hand side:

$$\frac{\partial T}{\partial \mathbf{q}} = \frac{1}{2}\sum_j \left[ I_j \frac{\partial(\dot{\theta}_j^2)}{\partial \mathbf{q}} \right] + \frac{1}{2}\sum_j \left[ m_j \frac{\partial(\dot{x}_j^2 + \dot{y}_j^2)}{\partial \mathbf{q}} \right]$$

$$= \frac{1}{2}\sum_j \left[ m_j \left( \frac{\partial(\dot{x}_j^2)}{\partial \mathbf{q}} + \frac{\partial(\dot{y}_j^2)}{\partial \mathbf{q}} \right) \right]$$

$$= \sum_j \left[ m_j \left( \dot{x}_j \frac{\partial \dot{x}_j}{\partial \mathbf{q}} + \dot{y}_j \frac{\partial \dot{y}_j}{\partial \mathbf{q}} \right) \right] \quad \text{(S21)}$$

$$= \sum_j \left[ \frac{\partial \dot{x}_j}{\partial \mathbf{q}} m_j \left( \frac{\partial x_j}{\partial \mathbf{q}} \right)^T \dot{\mathbf{q}} + \frac{\partial \dot{y}_j}{\partial \mathbf{q}} m_j \left( \frac{\partial y_j}{\partial \mathbf{q}} \right)^T \dot{\mathbf{q}} \right]$$

$$= \sum_j \left[ \frac{\partial \dot{x}_j}{\partial \mathbf{q}} m_j \left( \frac{\partial x_j}{\partial \mathbf{q}} \right)^T + \frac{\partial \dot{y}_j}{\partial \mathbf{q}} m_j \left( \frac{\partial y_j}{\partial \mathbf{q}} \right)^T \right] \dot{\mathbf{q}}$$

$$= \dot{\mathbf{J}}_{\mathbf{x}_{\text{seg}}}^T \mathbf{M} \mathbf{J}_{\mathbf{x}_{\text{seg}}} \dot{\mathbf{q}}.$$

For the fourth term on the left-hand side:

$$\frac{\partial V}{\partial \mathbf{q}} = \sum_j \left[ m_j g \frac{\partial y_j}{\partial \mathbf{q}} \right] = \mathbf{J}_y^T \mathbf{g}, \quad \text{(S22)}$$

where

$$\mathbf{J}_y = \frac{\partial \mathbf{y}}{\partial \mathbf{q}^T},$$

$$\mathbf{y} = [y_1, y_2, \ldots, y_9]^T, \quad \text{(S23)}$$

$$\mathbf{g} = [m_1 g, m_2 g, \ldots, m_9 g]^T.$$

For the generalized force vector $\mathbf{Q}$ on the right-hand side, according to the virtual work principle, we have

$$\begin{aligned}
\delta W &= \mathbf{Q}^T \delta \mathbf{q} \\
&= \tau_{\text{r,h}} \frac{\partial q_{\text{r,h}}}{\partial \mathbf{q}^T} \delta \mathbf{q} + \tau_{\text{r,k}} \frac{\partial q_{\text{r,k}}}{\partial \mathbf{q}^T} \delta \mathbf{q} + \tau_{\text{r,a}} \frac{\partial q_{\text{r,a}}}{\partial \mathbf{q}^T} \delta \mathbf{q} + \tau_{\text{l,h}} \frac{\partial q_{\text{l,h}}}{\partial \mathbf{q}^T} \delta \mathbf{q} + \tau_{\text{l,k}} \frac{\partial q_{\text{l,k}}}{\partial \mathbf{q}^T} \delta \mathbf{q} + \tau_{\text{l,a}} \frac{\partial q_{\text{l,a}}}{\partial \mathbf{q}^T} \delta \mathbf{q} \\
&\quad + \tau_{\text{r,e}} \frac{\partial q_{\text{r,e}}}{\partial \mathbf{q}^T} \delta \mathbf{q} + \tau_{\text{l,e}} \frac{\partial q_{\text{l,e}}}{\partial \mathbf{q}^T} \delta \mathbf{q} + \tau_{\text{r,int}} \left( \frac{\partial q_{\text{r,h}}}{\partial \mathbf{q}^T} - \frac{\partial q_{\text{r,e}}}{\partial \mathbf{q}^T} \right) \delta \mathbf{q} + \tau_{\text{l,int}} \left( \frac{\partial q_{\text{l,h}}}{\partial \mathbf{q}^T} - \frac{\partial q_{\text{l,e}}}{\partial \mathbf{q}^T} \right) \delta \mathbf{q} \\
&\quad + \sum_j \left[ F_{\text{r,heel,n}} \frac{\partial y_j}{\partial \mathbf{q}^T} \delta \mathbf{q} + F_{\text{r,toe,n}} \frac{\partial y_j}{\partial \mathbf{q}^T} \delta \mathbf{q} + F_{\text{l,heel,n}} \frac{\partial y_j}{\partial \mathbf{q}^T} \delta \mathbf{q} + F_{\text{l,toe,n}} \frac{\partial y_j}{\partial \mathbf{q}^T} \delta \mathbf{q} \right] \\
&\quad + \sum_j \left[ F_{\text{r,heel,f}} \frac{\partial y_j}{\partial \mathbf{q}^T} \delta \mathbf{q} + F_{\text{r,toe,f}} \frac{\partial y_j}{\partial \mathbf{q}^T} \delta \mathbf{q} + F_{\text{l,heel,f}} \frac{\partial y_j}{\partial \mathbf{q}^T} \delta \mathbf{q} + F_{\text{l,toe,f}} \frac{\partial y_j}{\partial \mathbf{q}^T} \delta \mathbf{q} \right] \\
&= \left( \boldsymbol{\tau}_{\text{joint}} + \boldsymbol{\tau}_{\text{exo}} + \boldsymbol{\tau}_{\text{int}} + \mathbf{J}_{\text{grf}}^T \mathbf{F}_{\text{grf}} \right)^T \delta \mathbf{q},
\end{aligned} \quad \text{(S24)}$$

where

$$\boldsymbol{\tau}_{\text{joint}} = \left[0,0,0,\tau_{\text{r,h}},\tau_{\text{r,k}},\tau_{\text{r,a}},0,\tau_{\text{l,h}},\tau_{\text{l,k}},\tau_{\text{l,a}},0\right]^{\text{T}},$$

$$\boldsymbol{\tau}_{\text{exo}} = \left[0,0,0,0,0,0,\tau_{\text{r,e}},0,0,0,\tau_{\text{l,e}}\right]^{\text{T}},$$

$$\boldsymbol{\tau}_{\text{int}} = \left[0,0,0,\tau_{\text{r,int}},0,0,-\tau_{\text{r,int}},\tau_{\text{l,int}},0,0,-\tau_{\text{l,int}}\right]^{\text{T}},$$

$$\mathbf{F}_{\text{grf}} = \left[F_{\text{r,heel,f}},F_{\text{r,toe,f}},F_{\text{l,heel,f}},F_{\text{l,toe,f}},F_{\text{r,heel,n}},F_{\text{r,toe,n}},F_{\text{l,heel,n}},F_{\text{l,toe,n}}\right]^{\text{T}}, \quad (S25)$$

$$\mathbf{J}_{\text{grf}} = \left[\mathbf{J}_{\text{f}}^{\text{T}},\mathbf{J}_{\text{n}}^{\text{T}}\right]^{\text{T}},$$

$$\mathbf{J}_{\text{f}} = \left[\frac{\partial \mathbf{x}_{\text{grf}}}{\partial \mathbf{q}^{\text{T}}}\right], \mathbf{J}_{\text{n}} = \left[\frac{\partial \mathbf{y}_{\text{grf}}}{\partial \mathbf{q}^{\text{T}}}\right].$$

$\mathbf{x}_{\text{grf}}$ and $\mathbf{y}_{\text{grf}}$ are the x- and y- coordinate vectors of the heels and toes with respect to $\{W\}$:

$$\mathbf{x}_{\text{grf}} = \left[x_{\text{r,heel}},x_{\text{r,toe}},x_{\text{l,heel}},x_{\text{r,toe}}\right]^{\text{T}},$$
$$\mathbf{y}_{\text{grf}} = \left[y_{\text{r,heel}},y_{\text{r,toe}},y_{\text{l,heel}},y_{\text{r,toe}}\right]^{\text{T}}. \quad (S26)$$

So the generalized force vector is

$$\mathbf{Q} = \boldsymbol{\tau}_{\text{joint}} + \boldsymbol{\tau}_{\text{exo}} + \boldsymbol{\tau}_{\text{int}} + \mathbf{J}_{\text{grf}}^{\text{T}} \mathbf{F}_{\text{grf}}. \quad (S27)$$

Finally, the EOM of the system can be written by combining all terms:

$$\mathbf{D}(\mathbf{q})\ddot{\mathbf{q}} + \mathbf{C}(\mathbf{q},\dot{\mathbf{q}})\dot{\mathbf{q}} + \mathbf{G}(\mathbf{q}) = \mathbf{Q}, \quad (S28)$$

where

$$\mathbf{D}(\mathbf{q}) = \mathbf{I} + \mathbf{J}_{\mathbf{x}_{\text{seg}}}^{\text{T}} \mathbf{M} \mathbf{J}_{\mathbf{x}_{\text{seg}}},$$
$$\mathbf{C}(\mathbf{q},\dot{\mathbf{q}}) = \mathbf{J}_{\mathbf{x}_{\text{seg}}}^{\text{T}} \mathbf{M} \dot{\mathbf{J}}_{\mathbf{x}_{\text{seg}}}, \quad (S29)$$
$$\mathbf{G}(\mathbf{q}) = \mathbf{J}_{\text{y}}^{\text{T}} \mathbf{g}.$$

# S3 Value of prescribed parameters

**Table S1.** Physiological parameters used in the muscle reflex model, muscle activation model, and the musculotendon contraction dynamic model

| Symbol | Description | Value | Unit |
| --- | --- | --- | --- |
| $F_{opt}^{TA}$ | Optimal muscle force of TA | 1759 | N |
| $F_{opt}^{SOL}$ | Optimal muscle force of SOL | 3549 | N |
| $F_{opt}^{GAS}$ | Optimal muscle force of GAS | 2342 | N |
| $F_{opt}^{FEM}$ | Optimal muscle force of FEM | 4530 | N |
| $F_{opt}^{HAM}$ | Optimal muscle force of HAM | 2594 | N |
| $F_{opt}^{GLU}$ | Optimal muscle force of GLU | 1944 | N |
| $F_{opt}^{ILI}$ | Optimal muscle force of ILI | 1759 | N |
| $l_{opt}^{TA}$ | Optimal muscle fiber length of TA | 0.098 | m |
| $l_{opt}^{SOL}$ | Optimal muscle fiber length of SOL | 0.05 | m |
| $l_{opt}^{GAS}$ | Optimal muscle fiber length of GAS | 0.06 | m |
| $l_{opt}^{FEM}$ | Optimal muscle fiber length of FEM | 0.087 | m |
| $l_{opt}^{HAM}$ | Optimal muscle fiber length of HAM | 0.109 | m |
| $l_{opt}^{GLU}$ | Optimal muscle fiber length of GLU | 0.147 | m |
| $l_{opt}^{ILI}$ | Optimal muscle fiber length of ILI | 0.1 | m |
| $\tau_{act}$ | Activation delay constant in the muscle activation model | 0.01 | s |
| $\tau_{deact}$ | Deactivation delay constant in the muscle activation model | 0.04 | s |
| $a_{min}$ | Minimum value of muscle activation | 0.01 | |
| $l_S^{T,TA}$ | Tendon slack length of TA | 0.223 | m |
| $l_S^{T,SOL}$ | Tendon slack length of SOL | 0.25 | m |
| $l_S^{T,GAS}$ | Tendon slack length of GAS | 0.39 | m |
| $l_S^{T,FEM}$ | Tendon slack length of FEM | 0.136 | m |
| $l_S^{T,HAM}$ | Tendon slack length of HAM | 0.31 | m |
| $l_S^{T,GLU}$ | Tendon slack length of GLU | 0.127 | m |
| $l_S^{T,ILI}$ | Tendon slack length of ILI | 0.163 | m |
| $\alpha_{opt}^{TA}$ | Pennation angle of TA when optimal force is generated | 5 | deg |
| $\alpha_{opt}^{SOL}$ | Pennation angle of SOL when optimal force is generated | 25 | deg |
| $\alpha_{opt}^{GAS}$ | Pennation angle of GAS when optimal force is generated | 17 | deg |
| $\alpha_{opt}^{FEM}$ | Pennation angle of FEM when optimal force is generated | 3 | deg |
| $\alpha_{opt}^{HAM}$ | Pennation angle of HAM when optimal force is generated | 0 | deg |
| $\alpha_{opt}^{GLU}$ | Pennation angle of GLU when optimal force is generated | 0 | deg |
| $\alpha_{opt}^{ILI}$ | Pennation angle of ILI when optimal force is generated | 8 | deg |
| $\beta$ | Proportionality coefficient of DE in the musculotendon contraction dynamics model | 0.1 | |

**Table S2.** Parameters used in the models of human-exoskeleton interaction and foot-ground contact

| Symbol | Description | Value | Unit |
|---|---|---|---|
| $k_{int}$ | Stiffness coefficient of the human-exoskeleton interaction model | 100 | N·m·deg$^{-1}$ |
| $d_{int}$ | Damping coefficient of the human-exoskeleton interaction model | 75 | N·m·s·deg$^{-1}$ |
| $k$ | Stiffness coefficient of the normal ground reaction force model | 160000 | N·m$^{-3/2}$ |
| $c$ | Dissipation coefficient of the normal ground reaction model | 1 | s·m$^{-1}$ |
| $\mu_s$ | Coefficient of static friction in the friction model | 0.9 | |
| $\mu_d$ | Coefficient of dynamic friction in the friction model | 0.6 | |
| $\mu_v$ | Coefficient of viscous friction in the friction model | 0.6 | s·m$^{-1}$ |
| $v_t$ | Transition velocity in the friction model | 0.15 | m·s$^{-1}$ |

**Table S3.** Parameters used in model optimization

| Symbol | Description | Value | Unit |
|---|---|---|---|
| $\psi$ | The desired distance from the model to the destination | 10 | M |
| $\omega_{vel}$ | Weight coefficient of $J_{vel}$ | 100 | |
| $\omega_{angle}$ | Weight coefficient of $J_{angle}$ | 0.1 | |
| $\omega_{GRF}$ | Weight coefficient of $J_{GRF}$ | 10 | |
| $\omega_{effort}$ | Weight coefficient of $J_{effort}$ | 0.1 | |
| $\hat{q}_{a,max}$ | Upper threshold of the ankle motion | 60 | deg |
| $\hat{q}_{a,min}$ | Lower threshold of the ankle motion | 60 | deg |
| $\hat{F}_{GRF}$ | Threshold of the normal ground reaction force | 1096.4 | N |
| $m$ | Overall mass of the model | 74.5314 | kg |
| $g$ | Gravitational acceleration | 9.794 | m·s$^{-2}$ |